\documentclass{article}

\pdfoutput=1
\PassOptionsToPackage{numbers, compress}{natbib}  
\usepackage[final]{neurips_2022}  

\usepackage{subcaption}
\usepackage{supertabular}
\usepackage{colortbl}
\usepackage[dvipsnames]{xcolor}
\usepackage[utf8]{inputenc} 
\usepackage[T1]{fontenc}    
\usepackage{microtype}      
\usepackage{xcolor}         
\usepackage{xspace}
\usepackage{amsmath,amssymb,amsfonts,dsfont,pifont,bm,bbm,mathrsfs,mathtools,nicefrac}
\usepackage{algorithm,algpseudocode,listings}
\usepackage{wrapfig}
\usepackage{booktabs,multirow,adjustbox,diagbox,threeparttable}
\definecolor{citeblue}{RGB}{48,111,186}
\usepackage[pagebackref=false,breaklinks=true,colorlinks=true,citecolor=citeblue,bookmarks=false]{hyperref}
\usepackage{cleveref}  
\usepackage{tikz}
\usepackage{pgfplots}
\pgfplotsset{compat=1.17}

\crefname{section}{Sec.}{Secs.}
\Crefname{section}{Section}{Sections}
\crefname{table}{Tab.}{Tabs.}
\Crefname{table}{Table}{Tables}
\crefname{figure}{Fig.}{Figs.}
\Crefname{figure}{Figure}{Figures}
\crefname{equation}{Eq.}{Eqs.}
\Crefname{equation}{Equation}{Equations}
\definecolor{gbypink}{rgb}{0.99, 0.91, 0.95} 

\definecolor{codegreen}{rgb}{0,0.6,0}
\definecolor{codegray}{rgb}{0.5,0.5,0.5}
\definecolor{codepurple}{rgb}{0.58,0,0.82}
\definecolor{backcolour}{rgb}{1.0,1.0,1.0}
\lstdefinestyle{mystyle}{
    backgroundcolor=\color{backcolour},
    commentstyle=\color{codegreen},
    keywordstyle=\color{magenta},
    numberstyle=\tiny\color{codegray},
    stringstyle=\color{codepurple},
    basicstyle=\ttfamily\scriptsize,
    breakatwhitespace=false,
    breaklines=true,
    captionpos=b,
    keepspaces=true,
    numbers=left,
    numbersep=5pt,
    showspaces=false,
    showstringspaces=false,
    showtabs=false,
    tabsize=2
}
\newcommand{\tocite}[1]{\textcolor{red}{[TO CITE]}}
\newcommand{\method}{FST\xspace}
\newcommand{\supp}{\textit{Supplementary Material}\xspace}

\title{Learning from Future: A Novel Self-Training Framework for Semantic Segmentation}

\author{
    Ye Du$^{1,2}$ \quad
    Yujun Shen$^3$ \quad
    Haochen Wang$^{4}$ \quad
    Jingjing Fei$^5$ \quad
    Wei Li$^5$ \\
    \textbf{Liwei Wu}$^5$ \quad
    \textbf{Rui Zhao}$^{5,6}$ \quad
    \textbf{Zehua Fu}$^{1,2}$ \quad
    \textbf{Qingjie Liu}$^{1,2}$\thanks{Corresponding Author}\\[3pt]
    $^1$ State Key Laboratory of Virtual Reality Technology and Systems,\ Beihang University\\
    $^2$ Hangzhou Innovation Institute,\ Beihang University\\
    $^3$ The Chinese University of Hong Kong \\
    $^4$ Institute of Automation, Chinese Academy of Sciences \quad
    $^5$ SenseTime Research\\
    $^6$ Qing Yuan Research Institute, Shanghai Jiao Tong University, Shanghai, China \\[3pt]
    \small{\texttt{\{duyee, zehua\_fu, qingjie.liu\}}@buaa.edu.cn}  \quad
    \small{\texttt{shenyujun0302@gmail.com}} \\
    \small{\texttt{wanghaochen2022@ia.ac.cn}}
    \quad 
    \small{\texttt{\{feijingjing1, liwei1, wuliwei, zhaorui\}@sensetime.com}}
}

\begin{document}

\maketitle

\begin{abstract}

Self-training has shown great potential in semi-supervised learning.
Its core idea is to use the model learned on labeled data to generate pseudo-labels for unlabeled samples, and in turn teach itself.
To obtain valid supervision, active attempts typically employ a momentum teacher for pseudo-label prediction yet observe the confirmation bias issue, where the incorrect predictions may provide wrong supervision signals and get accumulated in the training process.
The primary cause of such a drawback is that the prevailing self-training framework acts as guiding the current state with previous knowledge, because the teacher is updated with the past student only.
To alleviate this problem, we propose a novel self-training strategy, which allows the model to \textit{learn from the future}.
Concretely, at each training step, we first virtually optimize the student (\textit{i.e.}, caching the gradients without applying them to the model weights), then update the teacher with the virtual future student, and finally ask the teacher to produce pseudo-labels for the current student as the guidance.
In this way, we manage to improve the quality of pseudo-labels and thus boost the performance.
We also develop two variants of our \textit{future-self-training} (\method) framework through peeping at the future both deeply (\method-D) and widely (\method-W).
Taking the tasks of unsupervised domain adaptive semantic segmentation and semi-supervised semantic segmentation as the instances, we experimentally demonstrate the effectiveness and superiority of our approach under a wide range of settings.
Code is available at \href{https://github.com/usr922/FST}{https://github.com/usr922/FST}.

\end{abstract}

\section{Introduction}\label{sec:intro}

\begin{figure}[t]
    \centering
    \includegraphics[width=1.0\textwidth]{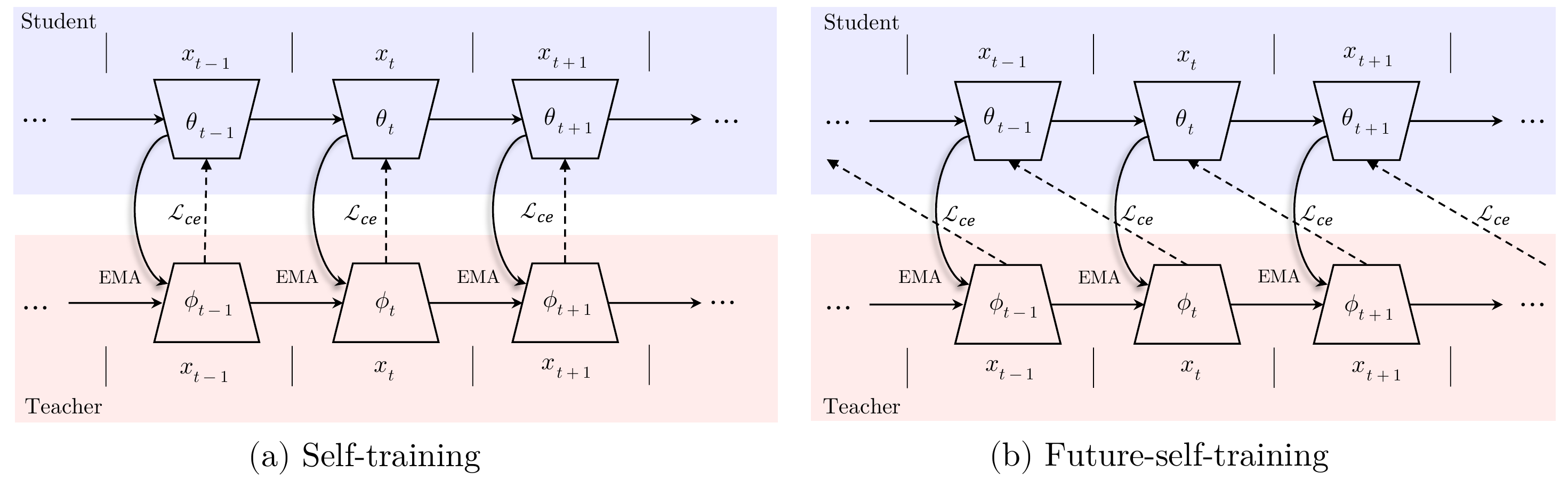}
    \vspace{-20pt}
    \caption{
        \textbf{Concept comparison} between self-training (ST) and our future-self-training (\method).
        (a) ST employs a teacher, which collects information from the \textit{past} states, to supervise the student.
        (b) Our \method derives a teacher at the \textit{future} moment and utilizes it to guide the current student.
    }
    \label{fig:st-vs-fst}
    \vspace{-2em}
\end{figure}

Improving the labeling efficiency of deep learning algorithms is vital in practice since acquiring high-quality annotations could consume great effort.
Self-training (ST) offers a promising solution to alleviate this issue by learning with limited labeled data and large-scale unlabeled data~\cite{van2020survey, hoyer2021daformer}.
The key thought is to learn a model on labeled samples and use it to generate pseudo-labels for unlabeled samples to teach the model itself.
In general, a teacher network that maintains an exponential moving average (EMA) of the student  (\textit{i.e.}, the model to learn) weights is used for pseudo-label prediction, as shown in \cref{fig:st-vs-fst}\textcolor{red}{a}.
Intuitively, such a training strategy relies on the \textit{previous} student states to supervise the current state, which amounts to using a poor model to guide a good one given the fact that a model tends to perform better along with the training process.
As a result, the confirmation bias issue~\cite{arazo2020pseudo, chen2022debiased} emerges from existing ST approaches, where the wrong supervision signals caused by those incorrect pseudo-labels get accumulated during training.

To break through the predicament of seeking supervision only from the past states, we propose \textit{future-self-training} (\method), which allows the model to learn from its \textit{future self}.
\cref{fig:st-vs-fst}\textcolor{red}{b} illustrates the concept diagram of our \method.
Compared to the conventional ST framework in \cref{fig:st-vs-fst}\textcolor{red}{a}, which employs the $t$-step teacher (\textit{i.e.}, updated with the student at moments $1, 2, \dots, t-1$) to guide the $t$-step student, \method presents a new training manner by urging the $t$-step student to learn from the $(t+1)$-step teacher.
However, at the start of the training step $t$, the $(t+1)$-step teacher is not available yet since it is dependent on the to-be-optimized $t$-step student.
To tackle this obstacle, we come up with a \textit{virtual updating} strategy.
Concretely, we first optimize the current student just like that in the traditional ST.
Differently, we do \textit{not} actually update the student weights but cache the gradients instead.
Such stashed gradients can be treated as the ``virtual future'' and help derive the $(t+1)$-step teacher.
Finally, the training of step $t$ borrows the pseudo-labels predicted by the latest teacher, and this time we apply the gradients to the student weights for real.

Recall that our motivation of encouraging the model to learn from the future is to help it acquire knowledge from an advanced teacher.
To this end, we put forward two variants based on our \method framework to make the teacher more capable.
On the one hand, we propose \method-D to investigate the future \textit{deeply}.
For this case, we ask the teacher to move forward for $K$ steps via virtual updating, thus the $t$-step student can be better supervised by the $(t+K)$-step teacher.
On the other hand, \method-W originates from the idea of model soups~\cite{wortsman2022model}, which reveals that the averaging weights of multiple fine-tuned models can improve the performance.
We hence propose to explore the future \textit{widely} with teachers developed from different training samples and expect the student to learn from all these $(t+1)$-step teachers simultaneously.

\begin{wrapfigure}[16]{R}{0.54\textwidth}
    \centering
    \vspace{-20pt}
    \includegraphics[width=0.58\textwidth]{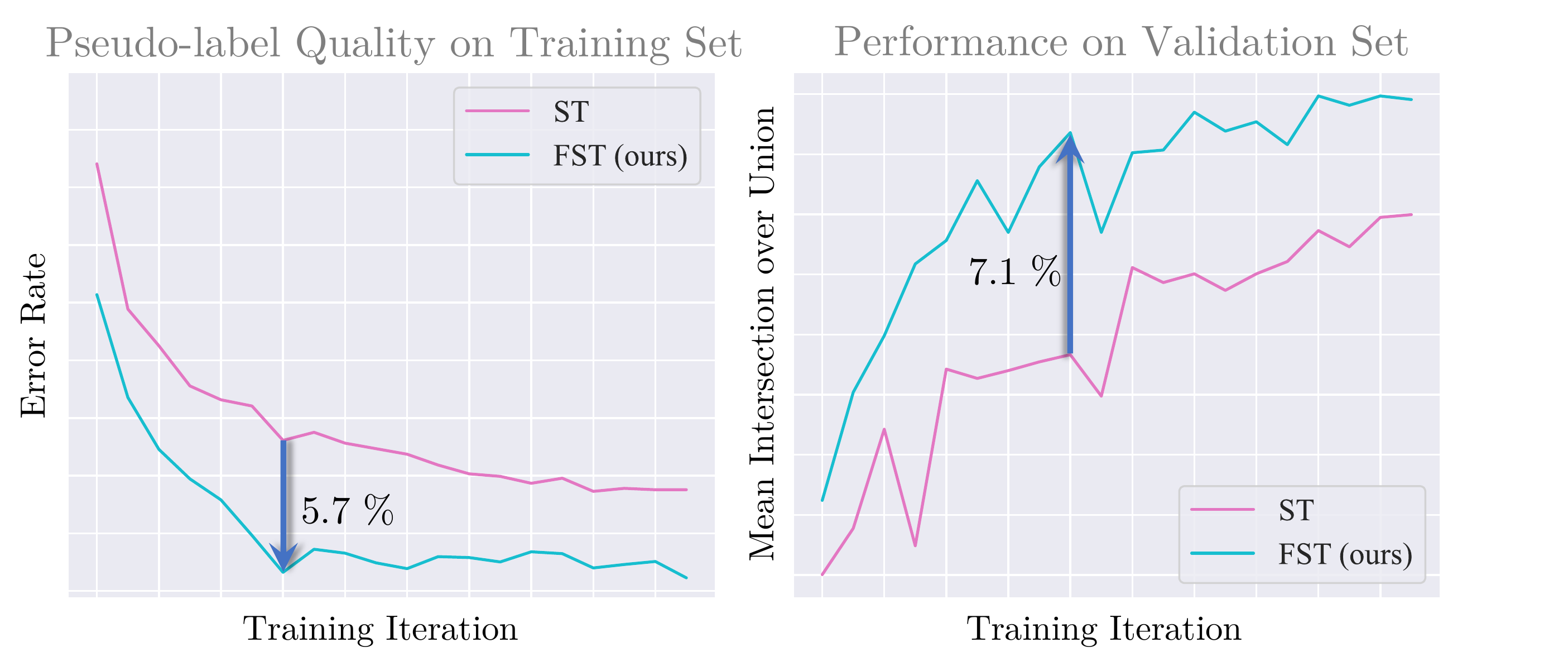}
    \vspace{-16pt}
    \caption{\label{fig-performance}
        \textbf{Performance comparison} between self-training (ST) and our future-self-training (\method), including the pseudo-label quality on unlabeled training samples (left) and the evaluation performance (right). 
        The comparison is conducted under the same number of updates of the student, which is the final model used for evaluation.
    }
\end{wrapfigure}

We evaluate our proposed \method on the tasks of both unsupervised domain adaptive (UDA) semantic segmentation and semi-supervised semantic segmentation.
The superiority of \method over the prevailing ST framework is summarized in \cref{fig-performance}, where our teacher model is capable of producing pseudo-labels with much higher quality and hence assists the student with a better performance.
This is because, along with the training process, the future states usually outperform the past states and thus can provide more accurate supervision, reducing the damage of confirmation bias.
Such a comparison validates our primary motive of learning from the future.
Furthermore, we observe consistent performance gain under a broad range of experimental settings (\textit{e.g.}, network architectures and datasets), demonstrating the effectiveness and generalizability of our approach.

\section{Related work}\label{sec:related}
\textbf{Domain adaptive semantic segmentation.} 
UDA semantic segmentation aims at transferring the knowledge from a labeled source domain to an unlabeled target domain, which is often viewed as a special semi-supervised learning problem.
Early methods for UDA segmentation focus on diminishing the distribution shift between the source and target domain at the input level \cite{hoffman2018cycada, sankaranarayanan2018learning, gong2019dlow}, the feature level \cite{tsai2018learning, chen2019progressive, chang2019all, li2021semantic}, or the output level \cite{tsai2018learning, vu2019advent, melas2021pixmatch}.
Over the years, adversarial learning \cite{goodfellow2014generative, ganin2016domain} has been the dominant approach to aligning the distributions.
However, the alignment-based methods may destroy the discrimination ability of features and cannot guarantee a small expected error on the target domain~\cite{zhang2021prototypical}.
In contrast, self-training \cite{amini2022self}, which is originated from semi-supervised learning (SSL) \cite{lee2013pseudo}, is introduced to directly minimize a proxy cross-entropy (CE) loss on the target domain.
By leveraging the model itself to generate pseudo-labels on unlabeled data, self-training together with tailored strategies such as consistency regularization \cite{zhou2020uncertainty, araslanov2021self}, cross-domain mixup \cite{tranheden2021dacs, zhou2021context}, contrastive learning \cite{kang2020pixel, liu2021domain, zhou2021domain, li2021semantic}, pseudo-label refine \cite{wang2021uncertainty, zhang2021prototypical, zheng2021rectifying}, auxiliary tasks \cite{vu2019dada, wang2021domain} and class balanced training \cite{li2022class} achieves excellent performance.
Recently, Hoyer et al. \cite{hoyer2021daformer} empirically proved that the transformer architecture \cite{xie2021segformer} is more robust to domain shift than CNN. 
They propose a transformer-based framework with three efficient training strategies in pursuit of milestone performance.

\textbf{Semi-supervised semantic segmentation.} 
Self-training is widely studied in SSL literature \cite{van2020survey}.
To facilitate the usage of unlabeled samples, Tarvainen et al. \cite{tarvainen2017mean} propose a mean teacher framework for consistency learning between a \textit{student} and a momentum updating \textit{teacher}.
This idea is extended later to semi-supervised semantic segmentation, which trains the student model with high-confident \textit{hard} pseudo-labels predicted by the teacher.
On this basis, extensive attempts improve semi-supervised semantic segmentation by CutMix augmentation \cite{DBLP:conf/bmvc/FrenchLAMF20}, class-balanced training \cite{zhou2021c3, hu2021semi, guan2022unbiased} and contrastive learning \cite{zhou2021c3, alonso2021semi, liu2021bootstrapping, wang2022semi}.
A closely relevant topic to self-training in SSL is consistency regularization, which believes that enforcing semantic or distribution consistency between various perturbations, such as image augmentation \cite{ke2020guided} and network perturbation \cite{zhang2020wcp}, can improve the robustness and generalization of the model.
In general, consistent regularization methods are used together with a ST framework.
We focus on improving the basic ST in this work.

\textbf{Nesterov's accelerated gradient descent.} 
A related idea to our work is Nesterov's accelerated gradient descent (NAG).
Originally proposed in \cite{Nes83} for solving convex programming problem, NAG is a first-order optimization method with a better convergence rate than gradient descent.
With the rise of deep learning, NAG is adopted as an alternative to momentum stochastic gradient descent (SGD) to optimize neural networks \cite{sutskever2013importance, dozat2016incorporating}.
It is intuitively considered to perform a look ahead gradient evaluation and then make a correction \cite{botev2017nesterov}.
Due to its solid theoretical explanations \cite{DBLP:conf/innovations/ZhuO17, DBLP:conf/icml/AssranR20} and remarkable performance, many works incorporate NAG with various tasks.
In \cite{DBLP:conf/iclr/LinS00H20}, Lin et al. adopt NAG into the area of adversarial attack, where they propose a Nesterov's iterative fast gradient sign method to improve the transfer ability of adversarial examples.
In \cite{yang2020federated}, Yang et al. explore the utilization of NAG in federal learning.
Different from NAG that pursues accelerated convergence, our work aims at building a stronger pseudo-label generator and improving the performance of traditional self-training.

\section{Method}\label{sec:method}
\begin{wrapfigure}[13]{R}{0.45\textwidth}
        \centering
        \vspace{-1.2em}
        \includegraphics[width=0.45\textwidth]{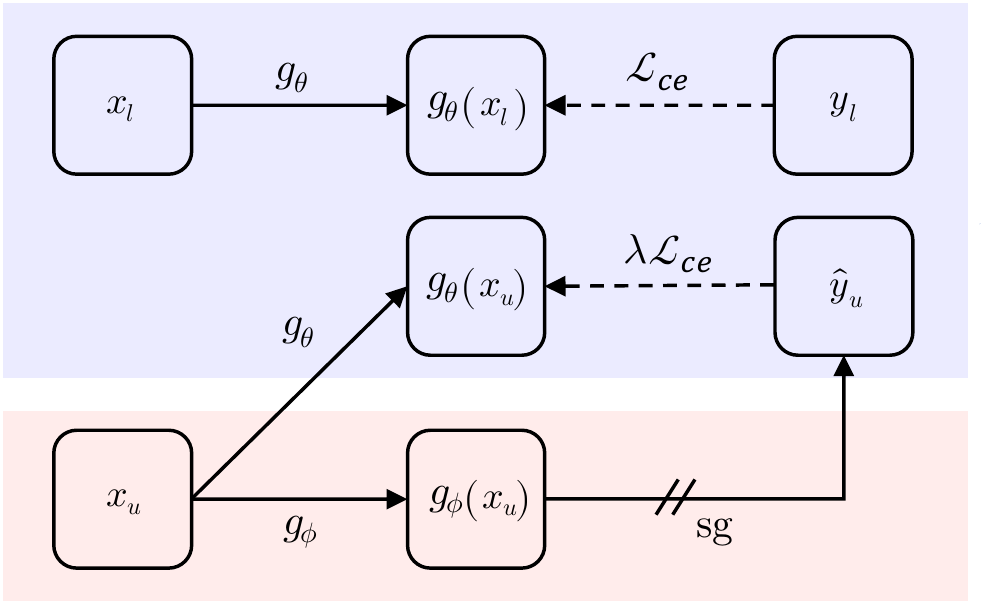}
        \caption{\label{fig1-mt}\textbf{Illustration} of the ST framework with a teacher $g_{\phi}$. ``sg'' means stop-gradient. }
\end{wrapfigure}
\subsection{Background}\label{sec:background}
Consider such a real-world scenario where we have access to a labeled segmentation dataset $\mathcal{D}_L = \{x_l, y_l\}_{l=1}^{n_l}$ from distribution $P$ and an unlabelled one $\mathcal{D}_U = \{x_u\}_{u=1}^{n_u}$ from unknown distribution $Q$.
We are required to build a semantic segmentation model using the combination of $\mathcal{D}_L$ and $\mathcal{D}_U$.
A general case is when $P\neq Q$, the problem falls into the category of UDA semantic segmentation. 
Otherwise, it is usually treated as a regular SSL task.

Self-training provides a unified solution and achieves state-of-the-art performance on both settings \cite{hoyer2021daformer, wang2022semi}.
One of the most common and widely used forms of self-training in semantic segmentation is a variant of mean teacher, which is shown in \cref{fig1-mt}.
Denote by $g_{\theta}$ the segmentation model required to be trained, and $\theta$ its parameters.
The mean teacher framework trains the \textit{student} $g_{\theta}$ on unlabeled data with pseudo-labels predicted by a momentum \textit{teacher} $g_{\phi}$, which has the same architecture to the student but with different parameters $\phi$.
Specifically, as the training progresses, the teacher evolves with the student by maintaining an EMA of student weights on each training iteration. 
This ensembling enables generating high quality predictions on unlabeled samples, and using them as training targets improves performance.
Formally, at each training step, the teacher is first updated and then predict pseudo-labels to train the student.
\begin{equation}
    \begin{aligned}
        \phi_{t+1} &= \mu \phi_{t} + (1 - \mu) \theta_{t}, \\
        \theta_{t+1}&= \theta_{t} - \gamma \nabla_{\theta} \left[ \mathcal{L}(g_{\theta_t}(x_l), y_l) + \lambda \mathcal{L}(g_{\theta_t}(x_u), \hat{y}_u|\phi_{t+1} ) \right],
    \end{aligned}
    \label{self-training}
\end{equation}
where $\mu$ is the momentum coefficient, $\gamma$ is the learning rate, and $\lambda$ is the dynamic re-weighting parameter to weigh the training of labeled and unlabeled data.
$\hat{y}_u$ denotes the pseudo-labels predicted by $\phi_{t+1}$, \textit{i.e.}, $\hat{y}_u=\arg\max g_{\phi_{t+1}}(x_u)$.
$\mathcal{L}$ is the pixel-wise cross-entropy training objective, which can be written as
\begin{equation}
    \begin{aligned}
        \mathcal{L}(x,y) = - \sum_{j=1}^{H\times W} \sum_{c=1}^C \mathbb{I}_{y^{j,c}=1} \text{log} g_{\theta}(x)^{j,c},
    \end{aligned}
    \label{ce-loss}
\end{equation}
where $H \times W$ is the input image size and $C$ is the total number of classes.

\textbf{Limitation of self-training.} Despite the remarkable performance, self-training suffers from the problem of confirmation bias. 
To be specific, the inherent noise in pseudo-labels could undesirably mislead the student training, which in return affects the pseudo-label prediction, and thereby results in noise accumulation.
Though a momentum updating strategy in the mean teacher framework improves tolerance with inaccurate pseudo-labels, this issue is still a bottleneck since the student still relies on learning from its own \textit{past} training states.

\subsection{Learning from future self}\label{sec:nesterov}
An intuitive observation shown in \cref{fig-performance} is that the performance of the student model generally improves during training, despite the noise in supervision.
From this perspective, a reasonable conjecture is, can we use model information from future moments to guide the current training iteration?
Motivated by this, we propose \textit{future-self-training} for facilitating the utilization of unlabeled data in semantic segmentation.
Concretely, at each training step, we propose to directly update the teacher model by the student weights from the next training moment.
To this end, a simple modification to \cref{self-training} is made as follows.
\begin{equation}
    \begin{aligned}
        \phi_{t+1} &= \mu \phi_{t} + (1 - \mu)\left(\theta_{t} -\gamma\nabla_{\theta} \left[\mathcal{L}(g_{\theta_t}(x_l),y_l)+\lambda\mathcal{L}(g_{\theta_{t}}(x_u),\hat{y}_u|\phi_{t})\right] \right), \\
        \theta_{t+1}&= \theta_{t} - \gamma \nabla_{\theta} \left[ \mathcal{L}(g_{\theta_t}(x_l), y_l) + \lambda \mathcal{L}(g_{\theta_t}(x_u), \hat{y}_u|\phi_{t+1} ) \right].
    \end{aligned}
    \label{naive-NST}
\end{equation}

Furthermore, it can be seen that \cref{naive-NST} only uses a virtual future state to update the teacher and ignores the current student weights $\theta_t$.
Our mission here is to establish a reliable pseudo-label generator (\textit{i.e.} a stronger teacher).
In terms of the ensembling effect of EMA, it is not necessary to discard $\theta_t$.
Therefore, an improved version of \method is proposed as follows.
\begin{equation}
    \begin{aligned}
      \phi'_{t+1} &= \mu \phi_{t}+(1-\mu)\theta_{t}, \\
      \phi_{t+1} &= \mu' \phi'_{t+1}  +(1-\mu')(\theta_{t}-\gamma \nabla_{\theta} [\mathcal{L}(g_{\theta_t}(x_l),y_l)+\lambda\mathcal{L}(g_{\theta_t}(x_u),\hat{y}_u| \phi'_{t+1} )]), 
      \\
        \theta_{t+1}&= \theta_{t} - \gamma \nabla_{\theta} \left[ \mathcal{L}(g_{\theta_t}(x_l), y_l) + \lambda \mathcal{L}(g_{\theta_t}(x_u), \hat{y}_u|\phi_{t+1} ) \right],
    \end{aligned}
    \label{imporved-NST}
\end{equation}
where a new momentum parameter $\mu'$ is introduced to distinguish the contribution of current and future model weights to teacher updates.
We provide pseudo-codes to further illustrate how we implement \cref{imporved-NST} in \supp.

\subsection{Exploring a deeper future}\label{sec:deeper}
We reiterate that the key insight of \method is to look ahead during training, which allows to mine more accurate supervision from future model states.
In experiments, we found that \cref{imporved-NST} exhibits only a slight improvement in performance (\cref{tab:ablation}), showing that this one-step future exploration strategy is insufficient.

Therefore, we further propose a looking ahead \textit{deeper} strategy to peek into deeper future student states.
To be specific, at each training step, we update the teacher not only with the student weights from the next moment, but also with those from deeper steps.
Formally, denote by $\widetilde{\phi}_t = \mu \phi_{t}+(1-\mu)\theta_{t}$ and $\widetilde{\theta}_{t}=\theta_t$ two agent variables firstly.
Then, we can use the co-evolving $\widetilde{\phi}_t$ and $\widetilde{\theta}_{t}$ for \textit{virtual updating} as follows.
\begin{equation}
    \begin{aligned}
        \widetilde{\theta}_{t+k+1} &=\widetilde{\theta}_{t+k}-\gamma \nabla_{\widetilde{\theta}} [\mathcal{L}(g_{ \widetilde{\theta}_{t+k}}(x_l),y_l)+\lambda\mathcal{L}( 
        g_{\widetilde{\theta}_{t+k}}(x_u),\hat{y}_u|\widetilde{\phi}_{t+k})], \\
        \widetilde{\phi}_{t+k+1} &= \mu'\widetilde{\phi}_{t+k} +(1-\mu')(\widetilde{\theta}_{t+k+1}),
    \end{aligned}
    \label{virtual-update}
\end{equation}
where $k=\{0, ..., K-1\}$ indexes the serial virtual steps for current training and $K$ is the total number of exploration steps.
Finally, we use the future information aware teacher $\widetilde{\phi}_{t+K}$ as the pseudo-label generator to supervise the current training. 
A simple reassignment and gradient descent update are applied to form the deeper version of \method, which is so called \method-D as shown below.
\begin{equation}
    \begin{aligned}
        \phi_{t+1} &= \widetilde{\phi}_{t+K}, \\
        \theta_{t+1} &=\theta_{t}-\gamma \nabla_{\theta} [\mathcal{L}(g_{\theta_t}(x_l),y_l)+\lambda\mathcal{L}(g_{\theta_t}(x_u),\hat{y}_u|\phi_{t+1})].
    \end{aligned}
    \label{NST-deeper}
\end{equation}

\subsection{Exploring a wider future}\label{sec:wider}
On the other hand, looking ahead \textit{wider} instead of \textit{deeper} is another intuitive way to enhance future exploration.
Inspired by the recent progress~\cite{wortsman2022model} that an ensemble of different model weights often shows excellent performance, we propose to first explore the next moment in different optimization directions and then use the average of them to update the teacher.
Concretely, we obtain different optimization directions by feeding \textit{different data batches} to the student model at each training moment.
Thus, a wider version of \method, \textit{i.e.}, \method-W, is presented as follows.
\begin{equation}
    \begin{aligned}
        \phi_{t+1} &= \mu' \{ \mu \phi_{t}+(1-\mu)\theta_{t} \} +(1-\mu')(\theta_{t}- \frac{1}{N}\sum_{i=1}^{N}\gamma\nabla_{\theta} [\mathcal{L}(g_{\theta_t}(x_l^{i}),y_l^{i})+\lambda\mathcal{L}(g_{\theta_t}(x_u^{i}),\hat{y}^{i}_u|\phi_{t})]), \\
        \theta_{t+1} &=\theta_{t}-\gamma \nabla_{\theta} [\mathcal{L}(g_{\theta_t}(x_l),y_l)+\lambda\mathcal{L}(g_{\theta_t}(x_u),\hat{y}_u|\phi_{t+1})],
    \end{aligned}
    \label{NST-wider}
\end{equation}
where $i$ indexes different samples and $N$ is parallel virtual exploration steps.

\cref{NST-wider} holds due to the fact that averaging the model weights is equivalent to averaging the gradients first and then updating the parameters by gradient descent.
It is worth noting that \method-D and \method-W are complementary that can be utilized together. 
However, this is beyond the scope of our work, and we leave this exploration to the future.

\section{Experiment}\label{sec:exp}
The experiment section is organized as follows.
First, we illustrate the experimental setup and implementation details in \cref{sec:exp-setup} and \cref{sec:exp-detail}.
Then, we evaluate the proposed \method and analyze the two variants in \cref{sec:eval}.
After that, we conduct extensive ablation studies to 
dissect our method in \cref{sec:exp_ablation}.
Finally, we compare our \method with existing state-of-the-art alternatives on both UDA and semi-supervised benchmarks in \cref{sec:exp-comparison}.

\subsection{Setup} \label{sec:exp-setup}
\textbf{Datasets and tasks.}
We evaluated our method on UDA and semi-supervised semantic segmentation.
In UDA segmentation, we use synthetic labeled images from GTAV \cite{richter2016playing_GTA} and SYNTHIA \cite{ros2016synthia} as the source domain and use real images from Cityscapes \cite{cordts2016cityscapes} as the target domain.
In addition, PASCAL VOC 2012 \cite{everingham2010pascal} is used for standard semi-supervised evaluation.
To simulate a semi-supervised setting, we randomly sample a portion \textit{(i.e.}, $1/4$, $1/8$, and $1/16$) of images together with corresponding segmentation masks from the training set as the labeled data and treat the rest as the unlabeled samples.

\textbf{Evaluation metric.}
Mean Intersection over Union (mIoU) is reported for evaluation. 
In SYNTHIA $\to$ Cityscapes UDA benchmark, $16$ and $13$ of the $19$ classes of Cityscapes are used to calculate mIoU, following the common practice~\cite{araslanov2021self, hoyer2021daformer}.

\textbf{Baselines.}
We first build strong baselines of the classical ST framework.
For UDA segmentation, we adopt the basic framework from \cite{tranheden2021dacs}, which contains a ClassMix augmentation.
Standard cross-entropy loss is calculated on both labeled and unlabeled data.
We use the efficient Encoder-Decoder structure for all semantic segmentation models, where the networks various in the structure of  encoders and decoders.
In the semi-supervised benchmark, we use the classical ST without other tricks as the baseline, because it has been proved to achieve competitive performance while maintaining simplicity~\cite{ke2020guided}.

\subsection{Implementation details}  \label{sec:exp-detail}
\textbf{Image augmentation}. 
The proposed \method and its baselines use the same image augmentation for fair comparison.
In UDA semantic segmentation, color jitter, Gaussian blur and ClassMix \cite{tranheden2021dacs} are used as the strong data augmentation for the unlabeled target domain, which follows the practice in \cite{hoyer2021daformer}.
In semi-supervised semantic segmentation, we use random flip and random crop, and the images are resized to $513 \times 513$ for both teacher and student.

\textbf{Network architecture}.
We use the DeepLabV2 \cite{chen2017deeplab} as the basic segmentation architecture for UDA segmentation, where the ASPP decoder only uses the dilation rates $6$ and $12$ following \cite{tsai2018learning}.
For Transformer-based networks, we adopt from \cite{hoyer2021daformer} and \cite{xiao2018unified} as the decoders.
In semi-supervised segmentation, we evaluate our method on the commonly used DeepLabV2 \cite{he2016deep}, DeepLabV3+ \cite{chen2018encoder} and PSPNet \cite{zhao2017pyramid_psp} with ResNet-101 \cite{he2016deep} as the backbone.

\textbf{Optimization}.
In UDA segmentation, the model is trained with an AdamW \cite{adam} optimizer, a learning rate of $6\times 10^{-5}$ for the encoder and $6\times 10^{-4}$ for the decoder, a weight decay of $0.01$, linear learning rate warmup with $1.5\text{k}$ iterations and linear decay afterwards.
We train the model on a batch of two $512\times 512$ random crops for a total of $40\text{k}$ iterations.
The momentum $u$ is set to $0.999$.
In semi-supervised segmentation, the model is trained with a SGD optimizer, a learning rate of $0.0001$ for the encoder and $0.001$ for the decoder, a weight decay of $0.0001$.
We train the model with $16$ labeled and $16$ unlabeled images per-batch for a total of $40$ epochs.

\subsection{Comparison with self-training} \label{sec:eval}

\begin{table}[t]
\centering
\small
\caption{\label{table:evaluation}
    \textbf{Comparison between ST and our \method}, where we explore the future with either (a) the same data batch as the current or (b) a different data batch from the current.
    ``SourceOnly'' means training the model with labeled data only, whose result is borrowed from \cite{hoyer2021daformer} as the reference.
    $4\times$ means using quadruple samples per mini-batch.
    All results are averaged over 3 random seeds.
}
\vspace{5pt}
\small
\scalebox{0.92}{
\begin{subtable}{.42\linewidth}
\begin{tabular}{lcc}
\toprule
Method       & mIoU           & $\Delta$  \\ \midrule
SourceOnly   & $34.3\pm 2.2$  &  -   \\
ST           & $56.3\pm 0.4$  & -    \\ 
-            & -              & -    \\
\midrule
Naive-\method    & $56.4\pm 0.4$  & $\uparrow 0.1$ \\
Improved-\method & $57.7\pm 0.6$  & $\uparrow 1.4$ \\
\method-W    & $56.8\pm 0.1$  & $\uparrow 0.5$ \\
\method-D    & $\cellcolor{gbypink} \bf{59.8\pm 0.1}$  & $\cellcolor{gbypink} \uparrow\bf{3.5}$ \\
\bottomrule
\end{tabular}
\caption{\label{table:same-batch}Future exploration with the \underline{same} data batch.}
\end{subtable} \hfill \hfill}
\scalebox{0.92}{
\begin{subtable}{.5\linewidth}
\begin{tabular}{lccc}
\toprule
Method        & Batch  & mIoU           & $\Delta$   \\ 
\midrule
SourceOnly    & $1\times$ & $34.3\pm 2.2$  &  -     \\
ST            & $1\times$  & $56.3\pm 0.4$  & -    \\ 
ST            & $4\times$  &$ 55.5 \pm 0.4$  & $\downarrow 0.8$    \\ 
\midrule
Naive-\method     & $1\times$  & $58.7\pm2.3$  & $\uparrow2.3$ \\
Improved-\method  & $1\times$  & $58.7\pm0.7$  & $\uparrow2.4$ \\
\method-W     & $1\times$  & $59.3\pm 0.5$  &  $\uparrow3.0$ \\
\method-D    & $1\times$  & $\cellcolor{gbypink} \bf{59.6\pm 1.4}$  & $\cellcolor{gbypink}\bf{\uparrow3.3}$ \\
\bottomrule
\end{tabular}
\caption{\label{table:diff-batch}Future exploration with a \underline{different} data batch.}
\end{subtable}}
\vspace{-15pt}
\end{table}

We first comprehensively compare our \method with classical ST to evaluate the effectiveness. 
The results are shown in \cref{table:evaluation}.
To simplify, we use GTAV as the labeled data and Cityscapes as the unlabeled data for evaluation.
All methods use the same experimental settings for fairness.

\begin{figure}[t]
    \centering
    \includegraphics[width=0.92\textwidth]{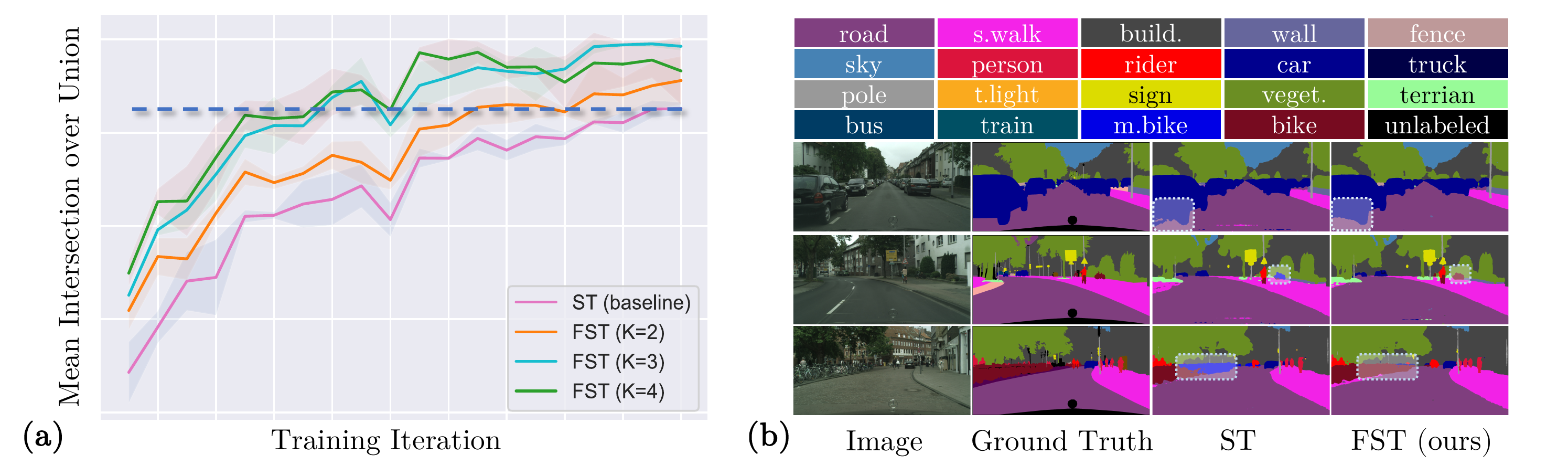}
    \vspace{-8pt}
    \caption{
        (a) \textbf{Performance curves} for ST and \method with various $K$ values. The comparison is conducted under the same number of updates of the student, which is the final model used for evaluation.
        (b) \textbf{Qualitative comparison} on Cityscapes~\cite{cordts2016cityscapes}, where dashed white boxes highlight the visual improvements.
    }
    \label{fig:analysis}
    \vspace{-13pt}
\end{figure}

\textbf{Quantitative analyses.}
We illustrate the improvements of Naive-\method (\cref{naive-NST}), Improved-\method (\cref{imporved-NST}), \method-D (\cref{virtual-update,NST-deeper}) and \method-W (\cref{NST-wider}) compared with classical ST (\cref{self-training}) in \cref{table:same-batch}.
These methods use the same batch of data for virtual forward at each step of future exploration.
As presented, Naive-\method only shows a negligible boost because the current student state is discarded without contributing to the teacher.
By revising it, the improved \method in \cref{NST-deeper},  which is a special case of \method-D when $K=1$, achieves an improvement of $1.4\%$ mIoU.
Further, \method-D (with $K=3$) clearly outperforms ST by a margin of $3.5\%$ mIoU, which benefits from the higher-quality pseudo-labels generated by a more reliable teacher as shown in \cref{fig-performance}.
In contrast, \method-W shows a slight improvement of only $0.5\%$ mIoU under the same data batch setting.
Thus, we prefer the deeper variant and adopt it as the basic approach in this paper, \textit{i.e.}, \method stands for \method-D unless specified.
We also analyse the effect of exploration steps (\textit{i.e.}, $K$) on the training process.
As suggested in \cref{fig:analysis}\textcolor{red}{a}, \method spends only about $1/3$ of the total training time to reach the performance level of ST.
Besides, we find that a larger $K$ can achieve higher mIoU at the beginning of the training process. 
When $K=4$, however, the performance in later training iterations drops and gets worse than $K=3$. 
We speculate that this is because the deeper exploration becomes unnecessary in the later training stage.
This interesting phenomenon indicates that an adaptive exploration mechanism may bring better results.

\textbf{Qualitative analyses.}
\cref{fig:analysis}\textcolor{red}{b} provides some qualitative comparisons, where our \method can correct some mistakes made by ST.
Taking the presented sample in the second row as an instance, ST struggles to distinguish between \textit{bicycle} and \textit{motorcycle}, while our \method successfully predicts it.
More visualization results and analyses can be found in \supp.

\textbf{Data batches for future exploration.} 
In \cref{sec:wider}, we derive \method-W, which uses different samples for future exploration in parallel.
\cref{table:same-batch} and \cref{table:diff-batch} compare the performance of using the same and different data batches.
Note that \method-W in \cref{table:same-batch} could produce slightly different mixed images for virtual forward, since we use ClassMix augmentation.
It is obvious that the parallel exploration with different samples performs better because the differences between models are important for ensembling.

\textbf{Generalization of popular architectures}.
To verify the generality under various advanced semantic segmentation models, we evaluate \method (the deeper variant) on two mainstream backbones (\textit{i.e.}, CNN and Transformer) with four commonly used segmentation decoders.
As presented in \cref{tab:ablation}, \method shows consistent performance improvement over classical ST, including DeepLab~\cite{chen2017deeplab}, PSPNet~\cite{zhao2017pyramid_psp} and UPerNet~\cite{xiao2018unified}.
Besides, \method shows significant improvements not only on supervised pretrained CNN~\cite{he2016deep} and Transformer backbones~\cite{Liu_2021_ICCV, xie2021segformer} but also on unsupervised pretrained BEiT~\cite{bao2021beit}.
Note that, the established ST baselines are strong, which even surpass many complex multi-stage methods (\textit{e.g.}, \cite{zhang2021prototypical}) proposed recently.
\method achieves $59.8\%$ mIoU using DeepLabV2 and ResNet-101. 
More comparisons between \method and existing CNN-based methods are provided in \supp.

\begin{table}[t]
\small
\centering
\caption{\label{tab:ablation}\textbf{Generalization} of \method across architectures. All results are averaged over 3 random seeds.}
\vspace{3.5pt}
\scalebox{0.82}{
\begin{subtable}{.38\linewidth}
    \begin{tabular}{lccc}
    \toprule
        Method & $K$ & mIoU & $\Delta$   \\ \midrule
        ST     & -    &  $55.0\pm 0.9$    &  -     \\
        \method    & $2$    &  $56.3\pm 1.0$    & $\uparrow 1.3$     \\
        \method    & $3$    &  $\cellcolor{gbypink}\bf{56.9\pm 0.5}$    & $\cellcolor{gbypink}\bf{\uparrow 1.9}$     \\
        \method    & $4$    &  $56.4\pm 0.9$    & $\uparrow 1.4$     \\ \bottomrule
    \end{tabular}
    \caption{\label{table:ablation-dlv2-r50}DeepLabV2 \cite{chen2017deeplab} \textit{w/} ResNet-50 \cite{he2016deep}.}
\end{subtable}\hfill}
\scalebox{0.82}{
\begin{subtable}{.38\linewidth}
    \begin{tabular}{lccc}
        \toprule
        Method  & $K$ & mIoU & $\Delta$   \\ \midrule
        ST      & -    &  $56.3\pm 0.4$    & -    \\
        \method     & $2$    &  $57.8\pm 1.3$    & $\uparrow 1.5$     \\
        \method     & $3$    &  $\cellcolor{gbypink}\bf{59.8\pm 0.1}$    & $\cellcolor{gbypink}\bf{\uparrow 3.5}$     \\
        \method     & $4$    &  $59.7\pm 0.8$    & $\uparrow 3.4$     \\ \bottomrule
    \end{tabular}
    \caption{\label{table:ablation-dlv2-r101}DeepLabV2 \cite{chen2017deeplab} \textit{w/} ResNet-101 \cite{he2016deep}.}
\end{subtable}\hfill}
\scalebox{0.82}{
\begin{subtable}{.38\linewidth}
    \begin{tabular}{lccc}
        \toprule
        Method & $K$ & mIoU & $\Delta$   \\ \midrule
        ST     & -    &  $56.3\pm 0.8$    & -     \\
        \method    & $2$    &  $58.1\pm 3.1$    & $\uparrow 1.8$     \\
        \method    & $3$    &  $58.5\pm 0.7$    & $\uparrow 2.2$     \\
        \method    & $4$    &  $\cellcolor{gbypink} \bf{58.8\pm 1.0}$    & $\cellcolor{gbypink} \bf{\uparrow 2.5}$     \\ \bottomrule
    \end{tabular}
    \caption{\label{table:ablation-psp-r101} PSPNet \cite{zhao2017pyramid_psp} \textit{w/} ResNet-101 \cite{he2016deep}.}
\end{subtable}}
\vfill
\scalebox{0.82}{
\begin{subtable}{.38\linewidth}
    \begin{tabular}{lccc}
        \toprule
        Method & $K$ & mIoU & $\Delta$  \\ \midrule
        ST     & -    &    $61.3\pm 0.7$    & -    \\
        \method    & $2$    &  $63.7\pm 2.0$    & $\uparrow 2.4$     \\
        \method   & $3$    &  $64.3\pm 2.3$    & $\uparrow 3.0$     \\
        \method   & $4$    &  $\cellcolor{gbypink}\bf{64.4\pm 2.0}$    & $\cellcolor{gbypink}\bf{\uparrow 3.1}$     \\ \bottomrule
    \end{tabular}
    \caption{\label{table:ablation-uper-swinb}UPerNet \cite{xiao2018unified} \textit{w/} Swin-B \cite{Liu_2021_ICCV}.}
\end{subtable}\hfill}
\scalebox{0.82}{
\begin{subtable}{.38\linewidth}
    \begin{tabular}{lccc}
        \toprule
        Method & $K$ & mIoU & $\Delta$  \\ \midrule
        ST     & -    &    $59.9\pm 2.0$    & -    \\
        \method    & $2$    &  $62.5\pm 1.2$    & $\uparrow 2.6$     \\
        \method    & $3$    &  $62.5\pm 1.9$    & $\uparrow 2.6$     \\
        \method    & $4$    &  $\cellcolor{gbypink} \bf{62.6\pm 1.8}$    & $\cellcolor{gbypink}\bf{\uparrow 2.7}$     \\ \bottomrule
    \end{tabular}
    \caption{\label{table:ablation-uper-beitb}UPerNet \cite{xiao2018unified} \textit{w/} BEiT-B~\cite{bao2021beit}.}
\end{subtable}\hfill}
\scalebox{0.82}{
\begin{subtable}{.38\linewidth}
    \begin{tabular}{lcccc}
        \toprule
        Method & $K$ & mIoU & $\Delta$  \\ \midrule
        ST     & -    &    $68.3\pm 0.5$    & -    \\
        \method    & $2$    &  $69.1\pm 0.3$    & $\uparrow 0.8$     \\
        \method    & $3$    &  $\cellcolor{gbypink}\bf{69.3\pm 0.3}$    & $\cellcolor{gbypink}\bf{\uparrow 1.0}$     \\
        \method    & $4$    &  $68.8\pm 0.9$    & $\uparrow 0.5$     \\ \bottomrule
    \end{tabular}
    \caption{\label{table:ablation-uper-mitb5}DAFormer \cite{hoyer2021daformer} \textit{w/} MiT-B5 \cite{xie2021segformer}.}
\end{subtable}}
\vspace{-18pt}
\end{table}

\subsection{Ablation studies}\label{sec:exp_ablation}

\textbf{Deeper or wider.} 
It can be seen from \cref{table:evaluation} that \method-D performs better than \method-W no matter using the same or different data batches for future exploration.
It is worth noting that under the setting of \cref{table:diff-batch}, the teacher model in \method sees more data per-iteration compared to the original ST, which has the effect of expanding the batch size in disguise.
To make a fair comparison, we also build a ST baseline with a larger batch size.
The results show that the performance gain of \method-W comes from the method itself instead of utilizing more data in each iteration.
In addition, we conduct ablations to compare \method-D and \method-W with each step (\textit{i.e.}, $K$ and $N$), which are shown in \cref{table:ablation-fst-d-r101-differdata} and \cref{table:ablation-fst-w-r101-differdata}.
Both variants use different data batches for future exploration since \method-W performs well only under this setting.
It can be concluded that \method-D performs better than \method-W, which is consistent with the conclusion in the above.
Besides, comparing \cref{table:ablation-dlv2-r101} and \cref{table:ablation-fst-d-r101-differdata}, we observe that using different data batches amplifies the performance jitter of each run.
We guess that this may be due to the diversity of data sampled for future exploration.

\textbf{Effect of the serial exploration steps $\boldsymbol{K}$}.
$K$ controls the number of virtual exploration steps in \method-D.
In \cref{tab:ablation}, we ablate $K$ on six settings, each evaluated over $3$ runs.
As presented, $K = 3$ shows steady improvements, while increasing it further brings negligible impact.
Thus, we recommend using $K=3$ as the basic practice.
We also ablate $K$ when using different data batches for exploration, which is presented in \cref{table:ablation-fst-d-r101-differdata}.

\textbf{Effect of the momentum $\boldsymbol{\mu'}$}. 
It is a common practice to set the momentum $\mu$ of EMA to a large value such as $0.999$ in self-training.
A separate momentum $\mu'$ that controls the contribution of future student states to the teacher is set in our \method.
We conduct ablation experiments to observe the effect of $\mu'$.
As shown in \cref{table:mu}, \method shows robustness against the change of $\mu'$. 
In our experiments, we set $\mu'=0.999$ as the default setting unless specifically stated, which equals the value of $\mu$.

\begin{table}[t]
\small
\centering
\caption{\label{tab:ablation-fst-w}
    \textbf{Analyses on the two variants of \method}, including \method-D (\cref{sec:deeper}) and \method-W (\cref{sec:wider}).
    All results are averaged over 3 random seeds.
}
\vspace{3.5pt}
\scalebox{0.92}{
\begin{subtable}{.52\linewidth}
    \begin{tabular}{llccc}
    \toprule
        Method & Backbone  & $K$ & mIoU & $\Delta$   \\ \midrule
        ST    & ResNet-101 & -    &  $56.3\pm 0.4$    &  -     \\
        \method-D  & ResNet-101  & $2$    &  $58.6\pm 0.4$    & $\uparrow 2.3$     \\
        \method-D & ResNet-101  & $3$    &  $59.6\pm 1.4$    & $\uparrow 3.3$     \\
        \method-D & ResNet-101   & $4$    &  $\cellcolor{gbypink}\bf{59.8\pm 2.0}$    & $\cellcolor{gbypink}\bf{\uparrow 3.5}$     \\ \bottomrule
    \end{tabular}
    \caption{\label{table:ablation-fst-d-r101-differdata} Effect of $K$ in \method-D.}
\end{subtable} \hfill }
\scalebox{0.92}{
\begin{subtable}{.52\linewidth}
    \begin{tabular}{llccc}
    \toprule
        Method & Backbone & $N$ & mIoU & $\Delta$   \\ \midrule
        ST     & ResNet-101 & -    &  $56.3\pm 0.4$    &  -     \\
        \method-W  &ResNet-101 & $2$    &  $58.5\pm 1.6$    & $\uparrow 2.2$     \\
        \method-W  & ResNet-101 & $3$    &  $\cellcolor{gbypink}\bf{59.3\pm 0.5}$    & $\cellcolor{gbypink}\bf{\uparrow 3.0}$     \\
        \method-W  & ResNet-101 & $4$    &  $58.6\pm 2.0$    & $\uparrow 2.3$     \\ \bottomrule
    \end{tabular}
    \caption{\label{table:ablation-fst-w-r101-differdata} Effect of $N$ in \method-W.}
\end{subtable}}
\vspace{-18pt}
\end{table}

\begin{table}[t]
\centering
\caption{\label{table:ablation_u_longer}
    (a) \textbf{Ablation study} on the hyper-parameter $\mu'$ (\cref{sec:nesterov}).
    (b) \textbf{Comparison with longer-training baselines.}
    All results are averaged over 3 random seeds.
}
\vspace{2pt}
\small
\scalebox{0.92}{
\begin{subtable}{.43\linewidth}
\centering
\begin{tabular}{lrcc}
\toprule
Method        & $\mu'$  & mIoU           & $\Delta$   \\ \midrule
ST            & -  & $56.3\pm 0.4$  & -    \\  
-             & -  & -              & -    \\ \midrule
\method            & $0.99$  &$58.8\pm 1.6$  & $\uparrow 2.5$    \\
\method     & $0.999$  & $59.8 \pm 0.1$  & $\uparrow3.5$ \\
\method  & $0.9999$  & $58.7 \pm 0.6$  & $\uparrow2.4$ \\
\method     & $0.99999$  & $\cellcolor{gbypink} \bf{59.9 \pm 0.9}$  &  $\cellcolor{gbypink} \bf{\uparrow3.6}$ \\
\bottomrule
\end{tabular}
\caption{\label{table:mu}Effect of $\mu'$ in \cref{virtual-update}.}
\end{subtable} \hfill}
\scalebox{0.92}{
\begin{subtable}{.6\linewidth}
\centering
\begin{tabular}{lcccc}
\toprule
Method &  Backbone     & Schedule  & mIoU           & $\Delta$   \\ \midrule
ST     &  ResNet-101       & $1\times$  & $56.3\pm 0.4$  & -    \\ 
ST     &  ResNet-101    & $4\times$  &$ 59.3 \pm 0.6$  & $\uparrow 3.0$    \\ 
\method  &   ResNet-101  & $1\times$  & $\cellcolor{gbypink}\bf{59.8 \pm 0.1}$  & $\cellcolor{gbypink}\bf{\uparrow3.5}$ \\ \midrule
ST     &  MiT-B5       & $1\times$  & $68.3\pm 0.5$  & -  \\
ST     &  MiT-B5       & $3\times$  & $68.3\pm 1.1$  & $\uparrow 0.0$  \\
\method     &  MiT-B5       & $1\times$  & $\cellcolor{gbypink}\bf{69.1\pm 0.3}$  & $\cellcolor{gbypink}\bf{\uparrow 0.8}$  \\
\bottomrule
\end{tabular}
\caption{\label{table:longerschedule}Comparison with longer training schedules.}
\end{subtable}}
\vspace{-2em}
\end{table}

\textbf{Effect of the parallel exploration steps $\boldsymbol{N}$.} 
In \method-W, $N$ controls the parallel exploration steps of the next training moment.
We conduct ablation experiments in \cref{table:ablation-fst-w-r101-differdata} to verify the influence of $N$.
As can be seen, $N=3$ performs well among the evaluated values, which is a similar observation to $K$ and implies that it is an acceptable choice in practice.

\textbf{Longer training schedules}.
We perform forward and backward propagation to obtain weights as the estimation of future student states.
This simple implementation linearly increases the training time over the number of exploration steps, \textit{i.e.}, $K$.
Note that in our method, the student is trained with the same training iterations as classical ST and does \textit{not} see more samples per-iteration, thereby the comparisons in \cref{table:evaluation,tab:ablation} are totally fair.
Even though, we establish stronger baselines with longer training schedules and compare them with our method.
We find that the performance of the longer scheduled ST baseline decreases heavily in the later training stages as the model is fitting the noise in pseudo-labels.
Besides, as shown in \cref{table:longerschedule}, the performance of longer training baselines still performs worse than our \method, which further proves the effectiveness of our method.

\subsection{Comparison with state-of-the-art alternatives}  \label{sec:exp-comparison}
In this subsection, we evaluate our \method with state-of-the-art approaches on the tasks of semi-supervised semantic segmentation and unsupervised domain adaptive semantic segmentation.

\textbf{Evaluation on semi-supervised segmentation.}
We first evaluate the proposed \method on traditional semi-supervised semantic segmentation.
As shown in \cref{tab:semi-voc}, we compare \method (with $K=3$) with ST on three partition protocols.
Equipped with three commonly used semantic segmentation networks, \textit{i.e.}, PSPNet~\cite{zhao2017pyramid_psp}, DeepLabV2~\cite{chen2017deeplab}, and DeepLabV3+~\cite{chen2018encoder}, \method consistently improves classical ST by considerable margins.
For instance, on the $1/8$ partition protocol, \method with DeepLabV3+ outperforms ST by $1.87\%$ mIoU, showing substantial improvement.
In short, \method demonstrates remarkable performance on the semi-supervised benchmark.
More comprehensive comparisons between our \method against state-of-the-art alternatives can be found in \supp.

\begin{table}[t]
\centering
\small
\caption{\label{tab:semi-voc}
    \textbf{Evaluation on the semi-supervised learning (SSL) setting} on PASCAL VOC 2012~\cite{everingham2010pascal}, where
    $1/16$, $1/8$, and $1/4$ stand for using $664$, $1323$, and $2646$ samples as the labeled set, respectively.}
\vspace{3pt}
\scalebox{0.90}{
\begin{tabular}{l|ccc|ccc|ccc}
\toprule
 & \multicolumn{3}{c|}{PSPNet~\cite{zhao2017pyramid_psp}} & \multicolumn{3}{c|}{DeepLabV2~\cite{chen2017deeplab}} & \multicolumn{3}{c}{DeepLabV3+~\cite{chen2018encoder}} \\ \cmidrule{2-10}
Method & $1/16$     & $1/8$    & $1/4$    &$1/16$      & $1/8$     & $1/4$     & $1/16$      & $1/8$      & $1/4$    \\ 
\midrule
ST     & $65.47$ & $72.24$  &  $75.47$ &  $68.45$ &  $72.54$ &     $76.21$    &  $73.31$&  $74.20$ & $77.78$  \\
\method (ours)    &  $68.35$ & $72.77$ & $75.90$ & $69.43$ & $73.18$ & $76.32$ & $73.88$ & $76.07$ &  $78.10$     \\ 
\midrule
$\Delta$    &  $2.88\uparrow $ &  $0.53\uparrow$ & $0.43\uparrow$ &  $0.98\uparrow$ &  $0.64\uparrow$ & $0.11\uparrow $ & $0.57\uparrow$ &        $1.87\uparrow$  &  $0.32\uparrow$ \\ 
\bottomrule
\end{tabular}}
\vspace{-12pt}
\end{table}

\textbf{Evaluation on unsupervised domain adaptive semantic segmentation.}
We then compare our \method with previous self-training based competitors on UDA benchmark.
The results are presented in \cref{tab:gta}.
Our \method is build upon DAFormer \cite{hoyer2021daformer}, which is the state-of-the-art method currently.
On GTAV $\to$ Cityscapes benchmark, our \method exceeds DAFormer by a considerable margin of $1.0\%$ mIoU and shows dominant performance in most categories.
Compared to the source only trained model, \method surpasses it by $23.2\%$ mIoU, indicating a further step toward practical applications.
The bottom part of \cref{tab:gta} shows the comparisons on SYNTHIA $\to$ Cityscapes benchmark.
\method exceeds DAFormer by $1.0\%$ mIoU on $16$ classes.
Besides, our \method achieves $68.5\%$ mIoU on $13$ classes, exceeding DAFormer by $1.1\%$.
In summary, our \method achieves new state-of-the-art performance on the UDA benchmark.

\begin{table}[t]
\centering
\caption{
    \textbf{Evaluation on the unsupervised domain adaptation (UDA) setting} on two benchmarks.
    Our results are averaged over $3$ random seeds.
}
\label{tab:gta}
\vspace{3pt}
\setlength{\tabcolsep}{2.3pt}
\scalebox{0.7}{
\begin{tabular}{l | ccccccccccccccccccc | c}
\toprule
Method &
\rotatebox{90}{Road} & \rotatebox{90}{S.walk} & \rotatebox{90}{Build.} & \rotatebox{90}{Wall} & \rotatebox{90}{Fence} & \rotatebox{90}{Pole} & \rotatebox{90}{T.light} & \rotatebox{90}{Sign} & \rotatebox{90}{Veget.} & \rotatebox{90}{Terrain} & \rotatebox{90}{Sky} & \rotatebox{90}{Person} & \rotatebox{90}{Rider} & \rotatebox{90}{Car} & \rotatebox{90}{Truck} & \rotatebox{90}{Bus} & \rotatebox{90}{Train} & \rotatebox{90}{M.bike} & \rotatebox{90}{Bike} & mIoU \\
\midrule
\multicolumn{21}{c}{GTAV~\cite{richter2016playing_GTA} $\to$ Cityscapes~\cite{cordts2016cityscapes}} \\
\midrule
SourceOnly
& $ 76.1 $ & $ 18.7 $ & $ 84.6 $ & $ 29.8 $ & $ 31.4   $ & $ 34.5 $ & $ 44.8  $ & $ 23.4 $ & $ 87.5 $ & $ 42.6 $ & $ 87.3	$ & $ 63.4   $ & $ 21.2  $ & $ 81.1 $ & $ 39.3 $ & $ 44.6 $ & $ 2.9 $ & $ 33.2 $ & $ 29.7 $ & $ 46.1 $ \\
ProDA~\cite{zhang2021prototypical} 
& $ 87.8 $ & $ 56.0 $ & $79.7 $ & $46.3 $ & $44.8 $ & $45.6 $ & $53.5 $ & $53.5 $ & \underline{$88.6$} & $ 45.2 $ & $82.1 $ & $70.7 $ & $39.2 $ & $88.8 $ & $45.5 $ & $59.4$ & $ 1.0 $ & $48.9$ & $ 56.4$ & $ 57.5$ \\  
CPSL~\cite{li2022class}
& $92.3 $ & $59.9 $ & $84.9 $ & $45.7 $ & $29.7 $ & $\bf{52.8} $ & $\bf{61.5} $ & $\bf{59.5} $ & $87.9 $ & $41.5 $ & $85.0 $ & $\bf{73.0} $ & $35.5 $ & $90.4 $ & $48.7 $ & $73.9 $ & $26.3 $ & $53.8 $ & $53.9 $ & $ 60.8$ \\
DAFormer~\cite{hoyer2021daformer}
& $ \bf{95.7} $ & $ \bf{70.2} $ & $ \bf{89.4} $ & \underline{$ 53.5 $} & $ \bf{48.1} $ & $ 49.6 $ & $55.8$ & \underline{$59.4$} & $ \bf{89.9} $ & \underline{$ 47.9 $} & \underline{$ 92.5 $} & $ 72.2 $ & \underline{$ 44.7 $} & \underline{$ 92.3 $} & \underline{$ 74.5 $} &\underline{$ 78.2 $} & \underline{$ 65.1 $} & \underline{$ 55.9 $} & \underline{$ 61.8 $} & \underline{$ 68.3 $} \\ 
\method (ours)
& \underline{$95.3$} & \underline{$67.7$} & \underline{$89.3$} & $\bf{55.5}$ & \underline{$47.1$} & \underline{$50.1$} & \underline{${57.2}$} & $58.6$ & $\bf{89.9}$ & $\bf{51.0}$ & $\bf{92.9}$ & \underline{${72.7}$} & $\bf{46.3}$ & $\bf{92.5}$ & $\bf{78.0}$ & $\bf{81.6}$ & $\bf{74.4}$ & $\bf{57.7}$ & $\bf{62.6}$ & $\bf{69.3}$ \\ 
\midrule
\multicolumn{21}{c}{SYNTHIA~\cite{ros2016synthia} $\to$ Cityscapes~\cite{cordts2016cityscapes}} \\
\midrule
SourceOnly
& $ 56.5 $ & $ 23.3   $ & $ 81.3    $ & $ 16.0 $ & $ 1.3   $ & $ 41.0 $ & $ 30.0  $ & $ 24.1 $ & $ 82.4 $ & $ - $ & $ 82.5	$ & $ 62.3   $ & $ 23.8  $ & $ 77.7 $ & $ - $ & $ 38.1 $ & $ - $ & $ 15.0 $ & $ 23.7 $ & $ 42.4 $  \\
ProDA~\cite{zhang2021prototypical}   
& \underline{$ 87.8 $} & \underline{$45.7$} & $84.6 $ & $37.1 $ & $0.6 $ & $44.0 $ & $54.6 $ & $37.0 $ & $\bf{88.1} $ & $-$ &$84.4 $ & $74.2 $ & $24.3 $ & \underline{$88.2 $} & $-$  & $51.1 $ & $-$ &$40.5 $ & $45.6 $ & $55.5 $ \\
CPSL~\cite{li2022class}
& $87.2$ & $43.9 $ & $85.5 $ & $33.6 $ & $0.3 $ & $47.7 $ & $ \bf{57.4} $ & $37.2 $ & \underline{${87.8}$} & $ - $ & $88.5 $ & $\bf{79.0} $ & $32.0 $ & $\bf{90.6} $ & $-$ & $49.4$ & $ -$ & $50.8 $ & $59.8 $ & $57.9$ \\ 
DAFormer~\cite{hoyer2021daformer}
& $ 84.5 $ & $ 40.7   $ & $ \bf{88.4}    $ & \underline{$ 41.5 $} & \underline{$ 6.5 $} & \underline{$ 50.0 $} & \underline{$ 55.0 $} & $\bf{54.6 }$ & $ 86.0 $ & $ - $ & \underline{$ 89.8 $} & $ 73.2   $ & $\bf{48.2}  $ & $ 87.2 $ & $ - $ & \underline{$ 53.2 $} & $ - $ & \underline{$ 53.9 $} & \underline{$ 61.7 $} & \underline{$ 60.9 $} \\ 
\method (ours)
& $\bf{88.3} $ & $ \bf{46.1}$ & \underline{$ 88.0$} & $\bf{41.7} $ & $\bf{7.3} $ & $ \bf{50.1} $ & $ 53.6 $ & \underline{$ 52.5 $} & $ 87.4 $ & $- $ & $ \bf{91.5} $ & \underline{$ 73.9 $} & \underline{$ 48.1 $} & $ 85.3 $ & $ - $ & $\bf{58.6} $ & $ - $ & $\bf{55.9}$ & $ \bf{63.4} $ & $ \bf{61.9}$ \\
\bottomrule
\end{tabular}}
\vspace{-10pt}
\end{table}

\section{Conclusion}\label{sec:conclusion}
In this paper, we present a future-self-training framework for semantic segmentation.
As an alternative to classical self-training, our approach mitigates the confirmation bias problem and achieves better performance on both UDA and semi-supervised benchmarks.
The key insight of our method is to mine a model's own future states as supervision for current training.
To this end, we propose two variants, namely \method-D and \method-W, to explore the future states deeply and widely.
Experiments on a wide range of settings demonstrate the effectiveness and generalizability of our methods.

\textbf{Discussion.} \label{sec:limitation}
The major drawback of this work is that our approach is time-consuming, since we need to forward a temporary model to acquire virtual future model states.
Although the number of updates to the student does \textit{not} increase, it is better to optimize the process of obtaining future model parameters.
To this end, an acceptable way is to maintain an ahead student model to provide information from future moments, which trades space for time.
Besides, our approach is \textit{general} that can be applied to other self-training frameworks such as FixMatch~\cite{sohn2020fixmatch} and other tasks such as semi-supervised image recognition~\cite{tarvainen2017mean}, object detection~\cite{Liu_2021_ICCV}, few-shot learning~\cite{su2020does}, and unsupervised representation learning~\cite{grill2020bootstrap}.
We will conduct further studies on these issues in the future.

{\small \label{sec:ref}
\bibliographystyle{abbrvnat}
\bibliography{ref}
}

\newpage
\renewcommand\thefigure{S\arabic{figure}}
\renewcommand\thetable{S\arabic{table}}  
\renewcommand\theequation{S\arabic{equation}}
\renewcommand\thealgorithm{S\arabic{algorithm}}
\setcounter{algorithm}{0}
\setcounter{equation}{0}
\setcounter{table}{0}
\setcounter{figure}{0}
\setcounter{section}{0}
\renewcommand\thesection{\Alph{section}}


\section*{\textit{Supplementary Material}}
%
The supplementary material is organized as follows. 
\cref{sec-sup:detail} shows more dataset and implementation details. 
\cref{sec:moreablations} provide more ablation studies of our \method, including the ablation on SYNTHIA $\to$ Cityscapes and evaluation of various segmentation decoders. 
\cref{sec-sup:comp-uda} and \cref{sec-sup:comp-semi} present more comparisons of our \method with state-of-the-art methods on both UDA and SSL benchmarks.
\cref{sec-sup:analyses} analyzes the training process of our method and shows more visualization comparisons with classical self-training.
\cref{sec-sup:discussion} discusses the social impact and potential negative impact of our work.
\cref{sec-sup:pseudo-code} shows the pseudo-code of our \method.

\section{More details} \label{sec-sup:detail}
\textbf{Dataset detail}.
GTAV~\cite{richter2016playing_GTA} contains $24,966$ labeled synthetic images with the size of $1914\times 1052$.
SYNTHIA~\cite{ros2016synthia} consists of $9,400$ labeled synthetic images with the size of $1280\times 760$.
Cityscapes has $2,975$ training and $500$ validation images with size of $2048\times 1024$.
PASCAL VOC 2012 \cite{everingham2010pascal} consists of $21$ classes with $1,464$, $1,449$, and $1,456$ images for the training, validation, and test set, respectively.
Following the common practice in semantic segmentation, we use
the augmented training set \cite{hariharan2011semantic} that consists of $10,582$ images for training.

\textbf{Implementation detail}.
We adopt a dynamic re-weighting approach from \cite{tranheden2021dacs} to weigh the labeled and unlabeled data, which takes the proportion of pixel-wise reliable predictions as the quality estimation of the pseudo-label:
\begin{equation}
    \begin{aligned}
    \lambda &= \frac{\sum_{j=1}^{H\times W} \mathbb{I}_{\max_c g_{\phi}(x_u)^{j}  > \tau}}{H \times W},
    \end{aligned}
\end{equation}
where $\tau$ is the confidence threshold and is set to $0.968$ for all experiments, $j$ indexes each pixel in $x_u$.

The ClassMix augmentation~\cite{tranheden2021dacs} randomly selects $1/2$ classes in the source image and paste their pixels onto the target image.
The error rate of the pseudo-label is calculated by
\begin{equation}
    \begin{aligned}
        \epsilon = 1 - \frac{1}{N \times C} \sum_{i=1}^{N}\sum_{c=1}^C \frac{\sum_{j=1}^{H\times W} \mathbb{I}_{\hat{y}_i^{j,c}=1;  y_i^{j,c}=1}}{\sum_{j=1}^{H\times W} \mathbb{I}_{y_i^{j,c}=1}}.
    \end{aligned}
\end{equation}
Following the common practice in UDA~\cite{hoyer2021daformer}, we resize the images to $1024 \times 512$ pixels for Cityscapes and to $1280 \times 720$ pixels for GTAV, then a random crop of size $512\times 512$ is used for training.
ImageNet pretrained weights are used to initialize the backbones.
The exception is the UPerNet with BEiT, which is initialized with the official self-supervised trained weights.
The UDA models are trained on $1$ Telsa A$100$ GPU, and the semi-supervised models are trained on $4$ Telsa V$100$ GPUs.
Our work is built on the MMSegmentation framework.

\section{More ablation}\label{sec:moreablations}
\textbf{Improvements on $\boldsymbol{13}$ classes.}
Previous works also compare the performance on $13$ classes (denoted by mIoU*), which discards three (\textit{i.e.}, \textit{wall}, \textit{fence} and \textit{pole}) of the $16$ classes in SYNTHIA $\to$ Cityscapes benchmark.
As shown in \cref{tab:sub-over13}, compared with previous state-of-the-art model DAFormer, our method exceeds it by $1.1\%$ mIoU.
\begin{table}[t]
\centering
\caption{
\textbf{Comparison}.
Comparison with state-of-the-art methods on SYNTHIA $\to$ Cityscapes UDA benchmark.
The mIoU and the mIoU* indicate we compute mean IoU over $16$ and $13$ categories, respectively. 
The results are averaged over $3$ random seeds.
}
\label{tab:sub-over13}
\vspace{10pt}
\setlength{\tabcolsep}{2.3pt}
\scalebox{0.71}{
\begin{tabular}{l | ccccccccccccccccccc | cc}
\toprule
Method &
\rotatebox{90}{Road} & \rotatebox{90}{S.walk} & \rotatebox{90}{Build.} & \rotatebox{90}{Wall*} & \rotatebox{90}{Fence*} & \rotatebox{90}{Pole*} & \rotatebox{90}{T.light} & \rotatebox{90}{Sign} & \rotatebox{90}{Veget.} & \rotatebox{90}{Terrain} & \rotatebox{90}{Sky} & \rotatebox{90}{Person} & \rotatebox{90}{Rider} & \rotatebox{90}{Car} & \rotatebox{90}{Truck} & \rotatebox{90}{Bus} & \rotatebox{90}{Train} & \rotatebox{90}{M.bike} & \rotatebox{90}{Bike} & mIoU & mIoU* \\
\midrule
SourceOnly
& $ 56.5 $ & $ 23.3   $ & $ 81.3    $ & $ 16.0 $ & $ 1.3   $ & $ 41.0 $ & $ 30.0  $ & $ 24.1 $ & $ 82.4 $ & $ - $ & $ 82.5	$ & $ 62.3   $ & $ 23.8  $ & $ 77.7 $ & $ - $ & $ 38.1 $ & $ - $ & $ 15.0 $ & $ 23.7 $ & $ 42.4 $ & $ 47.7$ \\
\midrule
CorDA~\cite{wang2021domain}
& $\bf{93.3} $ & $\bf{61.6} $ & $ 85.3 $ & $ 19.6$ & $ 5.1 $ & $37.8 $ & $36.6$ & $ 42.8$ & $ 84.9$ & $ - $ & \underline{$ 90.4$} & $ 69.7$ & $ 41.8 $ & $85.6 $ & $ - $ & $ 38.4 $ & $ - $ & $32.6 $ & $53.9 $ & $55.0 $ & $ 62.8$ \\
ProDA~\cite{zhang2021prototypical}   
& $87.8 $ & $45.7 $ & $84.6 $ & $37.1 $ & $0.6 $ & $44.0 $ & $54.6 $ & $37.0 $ & $\bf{88.1} $ & $-$ &$84.4 $ & $74.2 $ & $24.3 $ & \underline{$88.2 $} & $-$  & $51.1 $ & $-$ &$40.5 $ & $45.6 $ & $55.5 $ & $62.0$ \\
CPSL~\cite{li2022class}
& $87.2$ & $43.9 $ & $85.5 $ & $33.6 $ & $0.3 $ & $47.7 $ & $ \bf{57.4} $ & $37.2 $ & \underline{${87.8}$} & $ - $ & $88.5 $ & $\bf{79.0} $ & $32.0 $ & $\bf{90.6} $ & $-$ & $49.4$ & $ -$ & $50.8 $ & $59.8 $ & $57.9$ & $ 65.3$ \\ 
DAFormer~\cite{hoyer2021daformer}
& $ 84.5 $ & $ 40.7   $ & $ \bf{88.4}    $ & \underline{$ 41.5 $} & \underline{$ 6.5 $} & \underline{$ 50.0 $} & \underline{$ 55.0 $} & $\bf{54.6 }$ & $ 86.0 $ & $ - $ & $ 89.8 $ & $ 73.2   $ & $\bf{48.2}  $ & $ 87.2 $ & $ - $ & \underline{$ 53.2 $} & $ - $ & \underline{$ 53.9 $} & \underline{$ 61.7 $} & \underline{$ 60.9 $} & \underline{$ 67.4$} \\ 
\midrule
\method (ours)
& \underline{$88.3$} & \underline{$ 46.1$} & \underline{$ 88.0$} & $\bf{41.7} $ & $\bf{7.3} $ & $ \bf{50.1} $ & $ 53.6 $ & \underline{$ 52.5 $} & $ 87.4 $ & $- $ & $ \bf{91.5} $ & \underline{$ 73.9 $} & \underline{$ 48.1 $} & $ 85.3 $ & $ - $ & $\bf{58.6} $ & $ - $ & $\bf{55.9}$ & $ \bf{63.4} $ & $ \bf{61.9} $ & $\bf{68.5}$ \\
\bottomrule
\end{tabular}}
\vspace{-5pt}
\end{table}
\textbf{Ablation on SYNTHIA.}
We also provide ablation results on SYNTHIA $\to$ Cityscapes UDA benchmark and the results are shown in \cref{table:supablation3}.
In the main paper, we provide experiment results with $K=3$ to keep the same settings with the GTA $\to$ Cityscapes benchmark.
However, it can be seen that $K=2$ performs better in SYNTHIA $\to$ Cityscapes benchmark.
\begin{table}[t]
    \centering
    \small
     \caption{\label{table:supablation3}\textbf{Ablation}. Improvements on SYNTHIA $\to$ Cityscapes UDA Benchmark. Mean and SD are reported over $3$ random seeds.  The mIoU and the mIoU* indicate we compute mean IoU over $16$ and $13$ categories, respectively.}
     \vspace{10pt}
    \begin{tabular}{llccccc}
    \toprule
        Method & Backbone & $K$ & mIoU & $\Delta$ & mIoU* & $\Delta$* \\ \midrule
        ST     & MiT-B5 & -    &  $60.9$    &  -  & $67.4$ & -   \\
        \method    & MiT-B5 & $2$    &  $\cellcolor{gbypink}\bf{62.0\pm 0.9} $    & $\cellcolor{gbypink} \bf{\uparrow 1.1}$    & $\cellcolor{gbypink} \bf{68.8 \pm 1.1}$ & $\cellcolor{gbypink} \bf{\uparrow1.4}$ \\
        \method    & MiT-B5 & $3$    &  $61.9\pm 0.4$    & $\uparrow 1.0$ & $68.5 \pm 0.5$ & $\uparrow1.1$    \\
        \method    & MiT-B5 & $4$    &  $61.3\pm 1.1$    & $\uparrow 0.4$    & $68.0\pm1.4$ & $\uparrow0.6$ \\ \bottomrule
    \end{tabular}
\end{table}

\textbf{Ablation on decoder.}
We compare our \method with ST with various popular decoder architectures, including Atrous Spatial Pyramid Pooling (ASPP)~\cite{chen2017deeplab}, Pyramid Pooling Module (PPM)~\cite{zhao2017pyramid_psp}, PPM with Feature Pyramid Network (PPM + FPN)~\cite{xiao2018unified}, an MLP decoder~\cite{xie2021segformer}, and the decoder of DAFormer (SepASPP)~\cite{hoyer2021daformer}.
The MLP head fuses multi-level features and upsamples the feature map to predict the segmentation mask, which is designed for Transformer-based segmentation model~\cite{xie2021segformer}.
SepASPP is a multi-level context-aware feature fusion decoder which uses depth-wise separable convolutions to reduce over-fitting.
As shown in \cref{tab:ablation-decoder}, our method shows consistency improvements with these decoders.
\begin{table}[t]
     \centering
     \small
     \caption{\label{tab:ablation-decoder}\textbf{Ablation}. Ablation on popular segmentation decoders. Experiments are done on GTA $\to$ Cityscapes benchmark. Mean and SD are reported over $3$ random seeds.}
     \vspace{10pt}
    \begin{tabular}{lllcc}
    \toprule
        Method & Encoder  & Decoder      &   mIoU   & $\Delta$  \\ \midrule
        ST     & ResNet-101 & MLP~\cite{xie2021segformer}            &  $55.4\pm1.1$  & -     \\
        \method    & ResNet-101 & MLP~\cite{xie2021segformer}           &  $\cellcolor{gbypink} \bf{56.4\pm0.3}$  & $\cellcolor{gbypink}\bf{\uparrow1.0}$     \\ \midrule
        ST     & ResNet-101 & ASPP~\cite{chen2017deeplab}            &  $56.3\pm 0.4$  & -     \\
        \method    & ResNet-101 & ASPP~\cite{chen2017deeplab}            &  $\cellcolor{gbypink} \bf{59.8\pm0.1}$  & $\cellcolor{gbypink} \bf{\uparrow 3.5}$     \\ \midrule
        ST     & ResNet-101 & SepASPP~\cite{hoyer2021daformer}            &  $56.4\pm 0.4$  & -     \\
        \method    & ResNet-101 & SepASPP~\cite{hoyer2021daformer}            &  $\cellcolor{gbypink} \bf{57.6\pm0.4}$  & $\cellcolor{gbypink} \bf{\uparrow 1.2}$     \\ \midrule
        ST     & ResNet-101 & PPM~\cite{zhao2017pyramid_psp}           &  $56.3\pm0.8$  & -     \\
        \method    & ResNet-101 & PPM~\cite{zhao2017pyramid_psp}            &  $\cellcolor{gbypink} \bf{58.5\pm0.8}$  & $\cellcolor{gbypink} \bf{\uparrow2.2}$     \\
        \midrule
        ST     & ResNet-101 & PPM+FPN~\cite{xiao2018unified}           &  $56.6\pm0.9$  & -     \\
        \method    & ResNet-101 & PPM+FPN~\cite{xiao2018unified}           &  $\cellcolor{gbypink} \bf{60.1\pm0.3}$  & $\cellcolor{gbypink}\bf{\uparrow3.5}$ \\
    \bottomrule
    \end{tabular}
\end{table}

\section{More comparisons on UDA benchmark}  \label{sec-sup:comp-uda}
Most studies use CNN as the backbone.
In this section, we also compare category performance of our method with other state-of-the-art CNN-based methods.
As shown in \cref{tab:sup-udaperformance}, our \method with ResNet-101 achieves competitive performance among existing methods.
Note that, we report the performances of ProDA \cite{zhang2021prototypical} and CPSL \cite{li2022class} in \cref{tab:sup-udaperformance} \textit{without} knowledge distillation (which uses self-supervised trained models) for a fair comparison.
On the SYNTHIA $\to$ Cityscapes benchmark, we set $\mu'=0.9999$ for our \method.
As shown in \cref{tab:supsynthiaperformance}, our method also demonstrates competitive performance, which is slightly lower than CPSL, a class-balanced training approach that is orthogonal to our work.

\begin{table}[t]
\centering
\caption{
\textbf{Comparison}. Category performance comparison with state-of-the-art CNN-based methods on UDA benchmark. Methods use \textbf{ResNet-101}~\cite{he2016deep} as the backbone. The results are averaged over $3$ random seeds.
}
\vspace{10pt}
\setlength{\tabcolsep}{2.3pt}
\label{tab:sup-udaperformance}
\scalebox{0.71}{
\begin{tabular}{l | ccccccccccccccccccc | c}
\toprule
Method &
\rotatebox{90}{Road} & \rotatebox{90}{S.walk} & \rotatebox{90}{Build.} & \rotatebox{90}{Wall} & \rotatebox{90}{Fence} & \rotatebox{90}{Pole} & \rotatebox{90}{T.light} & \rotatebox{90}{Sign} & \rotatebox{90}{Veget.} & \rotatebox{90}{Terrain} & \rotatebox{90}{Sky} & \rotatebox{90}{Person} & \rotatebox{90}{Rider} & \rotatebox{90}{Car} & \rotatebox{90}{Truck} & \rotatebox{90}{Bus} & \rotatebox{90}{Train} & \rotatebox{90}{M.bike} & \rotatebox{90}{Bike} & mIoU \\
\midrule
AdaptSeg~\cite{tsai2018learning}
& $ 86.5 $ & $ 25.9    $ & $ 79.8     $ & $ 22.1 $ & $ 20.0  $ & $ 23.6 $ & $ 33.1  $ & $ 21.8 $ & $ 81.8       $ & $ 25.9    $ & $ 75.9 $ & $ 57.3   $ & $ 26.2  $ & $ 76.3 $ & $ 29.8  $ & $ 32.1 $ & $ 7.2   $ & $ 29.5      $ & $ 32.5 $ & $ 41.4 $ \\
ADVENT~\cite{vu2019advent}  
& $ 89.4 $ & $ 33.1    $ & $ 81.0     $ & $ 26.6 $ & $ 26.8  $ & $ 27.2 $ & $ 33.5  $ & $ 24.7 $ & $ 83.9       $ & $ 36.7    $ & $ 78.8 $ & $ 58.7   $ & $ 30.5  $ & $ 84.8 $ & $ 38.5  $ & $ 44.5 $ & $ 1.7   $ & $ 31.6      $ & $ 32.4 $ & $ 45.5 $\\
CBST~\cite{zou2018unsupervised}
& $ 91.8 $ & $ 53.5    $ & $ 80.5     $ & $ 32.7 $ & $ 21.0  $ & $ 34.0 $ & $ 28.9  $ & $ 20.4 $ & $ 83.9       $ & $ 34.2    $ & $ 80.9 $ & $ 53.1   $ & $ 24.0  $ & $ 82.7 $ & $ 30.3  $ & $ 35.9 $ & $ 16.0  $ & $ 25.9      $ & $ 42.8 $ & $ 45.9 $ \\
PCLA~\cite{kang2020pixel}
& $ 84.0 $ & $ 30.4    $ & $ 82.4     $ & $ 35.5 $ & $ 24.8  $ & $ 32.2 $ & $ 36.8  $ & $ 24.5 $ & $ 85.5       $ & $ 37.2    $ & $ 78.6 $ & $ 66.9   $ & $ 32.8  $ & $ 85.5 $ & $ 40.4  $ & $ 48.0 $ & $ 8.8  $ & $ 29.8      $ & $ 41.8 $ & $ 47.7$ \\
FADA~\cite{wang2020classes}
& $ 92.5 $ & $ 47.5    $ & $ 85.1     $ & $ 37.6 $ & $ 32.8  $ & $ 33.4 $ &$ 33.8 $ & $ 18.4 $ & $ 85.3       $ & $ 37.7    $ & $ 83.5 $ & $ 63.2   $ & \underline{$ 39.7 $} & $ 87.5 $ & $ 32.9  $ & $ 47.8 $ & $ 1.6   $ & $ 34.9      $ & $ 39.5 $ & $ 49.2 $ \\
MCS~\cite{chung2022maximizing}
& $ 92.6 $ & $ 54.0    $ & $ 85.4     $ & $ 35.0 $ & $ 26.0  $ & $ 32.4 $ & $ 41.2  $ & $ 29.7 $ & $ 85.1       $ & $ 40.9    $ & $ 85.4 $ & $ 62.6   $ & $ 34.7  $ & $ 85.7 $ & $ 35.6  $ & $ 50.8 $ & $ 2.4   $ & $ 31.0      $ & $ 34.0 $ & $ 49.7$ \\
CAG~\cite{zhang2019category}
& $ 90.4 $ & $ 51.6    $ & $ 83.8     $ & $ 34.2 $ & $ 27.8  $ & $ 38.4 $ & $ 25.3  $ & $ 48.4 $ & $ 85.4       $ & $ 38.2    $ & $ 78.1 $ & $ 58.6   $ & $ 34.6  $ & $ 84.7 $ & $ 21.9  $ & $ 42.7 $ & $ 41.1 $ & $ 29.3      $ & $ 37.2 $ & $ 50.2 $ \\
FDA~\cite{yang2020fda}
& $ 92.5 $ & $ 53.3    $ & $ 82.4     $ & $ 26.5 $ & $ 27.6  $ & $ 36.4 $ & $ 40.6  $ & $ 38.9 $ & $ 82.3       $ & $ 39.8    $ & $ 78.0 $ & $ 62.6   $ & $ 34.4  $ & $ 84.9 $ & $ 34.1  $ & $ 53.1 $ & $ 16.9  $ & $ 27.7      $ & $ 46.4 $ & $ 50.5 $\\
IAST~\cite{mei2020instance}    
& $ 93.8 $ & $ 57.8   $ & $ 85.1     $ & $ 39.5 $ & $ 26.7  $ & $ 26.2 $ & $ 43.1  $ & $ 34.7 $ & $ 84.9       $ & $ 32.9    $ & $ 88.0 $ & $ 62.6   $ & $ 29.0  $ & $ 87.3 $ & $ 39.2  $ & $ 49.6 $ & $ 23.2  $ & $ 34.7      $ & $ 39.6 $ & $ 51.5$ \\
DACS~\cite{tranheden2021dacs} 
& $89.9 $ & $39.7 $ & \underline{$87.9 $} & $30.7$ & $ 39.5 $ & $38.5$ & $ 46.4 $ & \underline{$52.8 $} & \underline{$88.0$} & $ 44.0 $ & $88.8 $ & $67.2 $ & $35.8$ & $ 84.5 $ & \underline{$45.7 $} & $50.2$ & $ 0.0 $ & $27.3 $ & $34.0 $ & $52.1$ \\
RCCR~\cite{zhou2021domain} 
& \underline{$ 93.7 $} & $ 60.4    $ & $ 86.5     $ & $ 41.0 $ & $ 32.0  $ & $ 37.3 $ & $ 38.7  $ & $ 38.6 $ & $ 87.2       $ & $ 43.0   $ & $ 85.5 $ & $ 65.4   $ & $ 35.1  $ & \underline{$ 88.3 $} & $ 41.8  $ & $ 51.6 $ & $ 0.0   $ & $ 38.0 $ & $ 52.1 $ & $ 53.5 $ \\
MetaCo~\cite{guo2021metacorrection}
& $ 92.8 $ & $ 58.1 $ & $ 86.2 $ & $39.7 $ & $33.1 $ & $36.3 $ & $42.0 $ & $38.6 $ & $85.5 $ & $37.8 $ & $87.6 $ & $62.8 $ & $31.7 $ & $84.8 $ & $35.7 $ & $50.3$ & $ 2.0$ & $ 36.8$ & $ 48.0$ & $ 52.1$ \\
CTF~\cite{ma2021coarse}
& $ 92.5 $ & $ 58.3 $ & $ 86.5 $ & $ 27.4 $ & $ 28.8 $ & $ 38.1 $ & $ 46.7 $ & $ 42.5 $ & $ 85.4 $ & $ 38.4 $ & $ \bf{91.8} $ & $ 66.4 $ & $ 37.0 $ & $ 87.8 $ & $ 40.7 $ & $ 52.4 $ & $ \bf{44.6} $ & $ 41.7 $ & $ \bf{59.0} $ & $ 56.1$ \\
CorDA~\cite{wang2021domain}
& $ 94.7 $ & \underline{$ 63.1 $} & $ 87.6 $ & $ 30.7 $ & $ \bf{40.6} $ & \underline{$ 40.2 $} & $ 47.8 $ & $ 51.6 $ & $ 87.6 $ & $ \bf{47.0} $ & \underline{$ 89.7 $} & $ 66.7 $ & $ 35.9 $ & $\bf{90.2}$ & $ \bf{48.9} $ & \underline{$ 57.5 $} & $ 0.0 $ & $ 39.8 $ & $ 56.0 $ & \underline{$ 56.6$}  \\
ProDA~\cite{zhang2021prototypical} 
& $91.5$ & $ 52.4 $ & $82.9 $ & \underline{$42.0 $} & $35.7 $ & $40.0$ & $ 44.4 $ & $43.3 $ & $87.0 $ & $43.8 $ & $79.5$ & $ 66.5$ & $ 31.4$ & $ 86.7 $ & $41.1 $ & $52.5 $ & $0.0 $ & $45.4 $ & $53.8 $ & $53.7$ \\  
CPSL~\cite{li2022class}
& $91.7$ & $ 52.9$ & $ 83.6$ & $ \bf{43.0} $ & $32.3 $ & $\bf{43.7}$ & $ \bf{51.3}$ & $ 42.8$ & $ 85.4$ & $ 37.6$ & $ 81.1$ & \underline{$ 69.5 $} & $ 30.0$ & $ 88.1$ & $ 44.1$ & $ \bf{59.9}$ & $ 24.9$ & $ \bf{47.2}$ & $ 48.4$ & $ 55.7$ \\
\midrule
\method (ours)
& $ \bf{95.0}  $ & $\bf{65.1}$ & $ \bf{88.4} $ & $ 40.1 $ & \underline{$ 36.8 $} & $ 38.0 $ & \underline{$ 50.2 $} & $ \bf{55.9} $ & $\bf{88.1} $ & \underline{$ 45.8 $} & $ 88.7 $ & $ \bf{70.1} $ & $\bf{45.0} $ & $ 87.4 $ & $ 45.3 $ & $ 54.8 $ & \underline{$ 37.2 $} & \underline{$ 45.6 $} & \underline{$ 58.9 $} & $\bf{ 59.8}$ \\
\bottomrule
\end{tabular}}
\vspace{-10pt}
\end{table}
\begin{table}[t]
\centering
\caption{\textbf{Comparison}.
Comparison with state-of-the-art CNN-based methods on SYNTHIA $\to$ Cityscapes UDA benchmark. Methods use \textbf{ResNet-101}~\cite{he2016deep} as the backbone.
The results are averaged over $3$ random seeds.
The mIoU and the mIoU* are calculated over $16$ and $13$ categories, respectively.}
\vspace{10pt}
\label{tab:supsynthiaperformance}
\setlength{\tabcolsep}{2.3pt}
\scalebox{0.72}{
\begin{tabular}{l | ccccccccccccccccccc | cc}
\toprule
Method &
\rotatebox{90}{Road} & \rotatebox{90}{S.walk} & \rotatebox{90}{Build.} & \rotatebox{90}{Wall*} & \rotatebox{90}{Fence*} & \rotatebox{90}{Pole*} & \rotatebox{90}{T.light} & \rotatebox{90}{Sign} & \rotatebox{90}{Veget.} & \rotatebox{90}{Terrain} & \rotatebox{90}{Sky} & \rotatebox{90}{Person} & \rotatebox{90}{Rider} & \rotatebox{90}{Car} & \rotatebox{90}{Truck} & \rotatebox{90}{Bus} & \rotatebox{90}{Train} & \rotatebox{90}{M.bike} & \rotatebox{90}{Bike} & mIoU & mIoU* \\
\midrule
AdaptSeg~\cite{tsai2018learning}
& $ 79.2 $ & $ 37.2    $ & $ 78.8     $ & $ -  $ & $ -   $ & $ - $ & $ 9.9   $ & $ 10.5  $ & $ 78.2 $ & $ - $ & $ 80.5 $ & $ 53.5   $ & $ 19.6  $ & $ 67.0 $ & $ - $ & $ 29.5 $ & $ - $ & $ 21.6 $ & $ 31.3 $ & $ - $ & $ 45.9$ \\
ADVENT~\cite{vu2019advent} 
& $ 85.6 $ & $ 42.2    $ & $ 79.7     $ & $ 8.7  $ & $ 0.4   $ & $ 25.9 $ & $ 5.4   $ & $ 8.1  $ & $ 80.4 $ & $ - $ & $ 84.1 $ & $ 57.9   $ & $ 23.8  $ & $ 73.3 $ & $ - $ & $ 36.4 $ & $ - $ & $ 14.2 $ & $ 33.0 $ & $ 41.2 $ & $ 48.0 $ \\

CBST~\cite{zou2018unsupervised}   
& $ 68.0 $ & $ 29.9    $ & $ 76.3     $ & $ 10.8 $ & $ 1.4   $ & $ 33.9 $ & $ 22.8  $ & $ 29.5 $ & $ 77.6 $ & $ - $ & $ 78.3 $ & $ 60.6   $ & $ 28.3  $ & $ 81.6 $ & $ - $ & $ 23.5 $ & $ - $ & $ 18.8 $ & $ 39.8 $ & $ 42.6 $ & $ 48.9$ \\
FDA~\cite{yang2020fda}
& $ 79.3 $ & $ 35.0    $ & $ 73.2     $ & $ -  $ & $ -   $ & $ - $ & $ 19.9  $ & $ 24.0 $ & $ 61.7 $ & $ - $ & $ 82.6 $ & $ 61.4   $ & $ 31.1  $ & $ 83.9 $ & $ - $ & $ 40.8 $ & $ - $ & $ 38.4 $ & $ 51.1 $ & $ - $ & $ 52.5 $\\
FADA~\cite{wang2020classes}  
& $ 84.5 $ & $ 40.1    $ & $ 83.1     $ & $ 4.8  $ & $ 0.0   $ & $ 34.3 $ & $ 20.1  $ & $ 27.2 $ & $ 84.8 $ & $ - $ & $ 84.0 $ & $ 53.5   $ & $ 22.6  $ & $ 85.4 $ & $ - $ & $ 43.7 $ & $ - $ & $ 26.8 $ & $ 27.8 $ & $ 45.2 $ & $ 52.5$ \\
MCS~\cite{chung2022maximizing}
& \underline{$ 88.3 $} & $ \bf{47.3}    $ & $ 80.1     $ & $ -  $ & $ -   $ & $ - $ & $ 21.6   $ & $ 20.2  $ & $ 79.6 $ & $ - $ & $ 82.1 $ & $ 59.0   $ & $ 28.2  $ & $ 82.0 $ & $ - $ & $ 39.2 $ & $ - $ & $ 17.3 $ & $ 46.7 $ & $ - $ & $ 53.2$ \\
PyCDA~\cite{lian2019constructing} 
& $ 75.5 $ & $ 30.9    $ & $ 83.3     $ & $ 20.8 $ & $ 0.7   $ & $ 32.7 $ & $ 27.3  $ & $ 33.5 $ & $ 84.7 $ & $ - $ & $ 85.0 $ & $ 64.1   $ & $ 25.4  $ & $ 85.0 $ & $ - $ & $ 45.2 $ & $ - $ & $ 21.2 $ & $ 32.0  $ & $ 46.7 $ & $ 53.3 $ \\
PLCA~\cite{kang2020pixel} 
& $ 82.6 $ & $ 29.0    $ & $ 81.0     $ & $ 11.2 $ & $ 0.2   $ & $ 33.6 $ & $ 24.9  $ & $ 18.3 $ & $ 82.8 $ & $ - $ & $ 82.3 $ & $ 62.1   $ & $ 26.5  $ & $ 85.6 $ & $ - $ & $ 48.9 $ & $ - $ & $ 26.8 $ & $ 52.2  $ & $ 46.8 $ & $ 54.0 $ \\
RCCR~\cite{zhou2021domain} 
& $ 79.4 $ & $ 45.3    $ & $ 83.3     $ & $ -  $ & $ -   $ & $ - $ & $ 24.7 $ & $ 29.6  $ & $ 68.9 $ & $ - $ & \underline{$ 87.5 $} & $ 61.1   $ & \underline{$ 33.8 $} & $ 87.0 $ & $ - $ & \underline{$ 51.0 $} & $ - $ & $ 32.1 $ & $ 52.1 $ & $ - $ & $ 56.8$ \\
IAST~\cite{mei2020instance}   
& $ 81.9 $ & $ 41.5    $ & $ 83.3     $ & $ 17.7 $ & $\bf{ 4.6 }  $ & $ 32.3 $ & $ 30.9  $ & $ 28.8 $ & $ 83.4 $ & $ - $ & $ 85.0 $ & $ 65.5   $ & $ 30.8  $ & $ 86.5 $ & $ - $ & $ 38.2 $ & $ - $ & $ 33.1 $ & $ 52.7 $ & $ 49.8 $ & $ 57.0$ \\
SAC~\cite{araslanov2021self}   
& $ \bf{89.3} $ & \underline{$47.2$} & $\bf{85.5} $ & $ 26.5 $ & $ 1.3 $ & $ \bf{43.0} $ & $ 45.5  $ & $ 32.0 $ & $ \bf{87.1} $ & $ - $ & $ \bf{89.3} $ & $ 63.6   $ & $ 25.4  $ & $ 86.9 $ & $ - $ & $ 35.6 $ & $ - $ & $ 30.4 $ & $ 53.0 $ & $ 52.6 $ & $ 59.3 $\\
ProDA~\cite{zhang2021prototypical}   
& $ 87.1 $ & $ 44.0   $ & $ 83.2    $ & $ \bf{26.9} $ & $ 0.7   $ & $ 42.0 $ & $ 45.8  $ & \underline{$ 34.2 $} & \underline{$ 86.7 $} & $ - $ & $ 81.3	$ & $ 68.4   $ & $ 22.1  $ & \underline{$ 87.7 $} & $ - $ & $ 50.0 $ & $ - $ & $ 31.4 $ & $ 38.6 $ & $ 51.9 $ & $ 58.5$ \\
CPSL~\cite{li2022class}
& $ 87.3 $ & $ 44.4   $ & \underline{$ 83.8 $} & \underline{$ 25.0 $} & $ 0.4   $ & \underline{$ 42.9 $} & $ \bf{47.5}  $ & $ 32.4 $ & $ 86.5 $ & $ - $ & $ 83.3 $ & $\bf{ 69.6}   $ & $ 29.1  $ & $ \bf{89.4} $ & $ - $ & $ \bf{52.1} $ & $ - $ & \underline{$ 42.6 $} & \underline{$ 54.1 $} & $ \bf{54.4} $ & $\bf{61.7}$ \\ 
\midrule
\method (ours)
& $68.5 $ & $28.9 $ & $\bf{85.5} $ & $21.1 $ & \underline{$3.3$}  & $40.4 $ & \underline{$46.3 $} & $ \bf{53.0} $ & $77.6 $ & $ - $ & $85.3 $ & \underline{$69.5 $} & $\bf{42.4} $ & $87.0 $ & $- $ & $48.5 $ & $- $ & $\bf{46.4} $ & $\bf{60.0} $ & \underline{$54.0 $} & \underline{$61.5$}
 \\
\bottomrule
\end{tabular}}
\vspace{-5pt}
\end{table}

\section{More comparisons on SSL benchmark} \label{sec-sup:comp-semi}
We compare our \method with previous state-of-the-art semi-supervised semantic segmentation frameworks, including CCT~\cite{ouali2020semi}, GCT~\cite{ke2020guided} and CPS~\cite{chen2021semi}.
These frameworks do \textit{not} use CutMix Augmentation~\cite{DBLP:conf/bmvc/FrenchLAMF20} for fair comparisons.
Experiments are conducted on both the PASCAL VOC 2012 and Cityscapes, with $1/16$, $1/8$ and $1/4$ samples as the labeled data.
The comparisons are shown in \cref{tab:semi-comparison}.
Note that some works such as AEL~\cite{hu2021semi} are not included here, since we compare our \method with the basic SSL frameworks.
However, AEL focuses on the long tail problem under a ST framework, which is orthogonal to our work.
On PASCAL VOC 2012, our \method achieves the best performance among these SSL frameworks.
On Cityscapes, our method exceeds CCT and GCT by large margins.
Compared to CPS, our \method also achieves competitive results.
Our \method uses minimal data augmentations, thus its performance could be further boosted by advanced augmentation strategies.
These results show the effectiveness of the proposed \method on the traditional SSL benchmark.
\begin{table}[t]
\centering
\small
\caption{\label{tab:semi-comparison}\textbf{Comparison}. Comparison with state-of-the-art semi-supervised semantic segmentation methods on the validation set. We use \method-D with $K=3$ and $\dagger$ means results reported by \cite{wang2022semi}.}
\vspace{10pt}
\begin{subtable}{0.45\textwidth}
\centering
\small
\begin{tabular}{lccc}
\toprule
Method     & $1/16$  & $1/8$   & $1/4$   \\ \midrule
SupOnly$^\dagger$    & $67.87$ & $71.55$ & $75.80$  \\ \midrule
CutMix$^\dagger$~\cite{DBLP:conf/bmvc/FrenchLAMF20}    & $71.66$ & $75.51$ & $77.33$ \\
CCT~\cite{ouali2020semi}        & $71.86$ & $73.68$ & $76.51$  \\
GCT~\cite{ke2020guided}        & $70.90$ & $73.29$ & $76.66$  \\
CPS~\cite{chen2021semi}        & \underline{$72.18$} & \underline{$75.83$} & \underline{$77.55$} \\ \midrule
\method (ours) & $\bf{73.88}$ & $\bf{76.0}7$ & $\bf{78.10}$ \\ 
\bottomrule
\end{tabular}
\caption{PASCAL VOC 2012~\cite{everingham2010pascal}.}
\end{subtable}
\begin{subtable}{0.5\textwidth}
\centering
\begin{tabular}{lccc}
\toprule

Method     & $1/16$  & $1/8$   & $1/4$   \\ \midrule
SupOnly$^\dagger$    & $65.74$ & $72.53$ & $74.43$  \\ \midrule
CutMix$^\dagger$~\cite{DBLP:conf/bmvc/FrenchLAMF20}     & $67.06$ & $71.83$ & $76.36$ \\
CCT~\cite{ouali2020semi}    & $69.32$ & $74.12$ & $75.99$  \\ 
GCT~\cite{ke2020guided}     & $66.75$ & $72.66$ & $76.11$  \\ 
CPS~\cite{chen2021semi}     & \underline{$70.50$} & $\bf{75.71}$ & $\bf{77.41}$ \\ \midrule
\method (ours) & $\bf{71.03}$ & \underline{$75.36$} & \underline{$76.61$} \\ 
\bottomrule
\end{tabular}
\caption{Cityscapes~\cite{cordts2016cityscapes}.}
\end{subtable}
\end{table}

\section{More analyses} \label{sec-sup:analyses}
\begin{figure}[t]
    \centering
    \includegraphics[width=.94\textwidth]{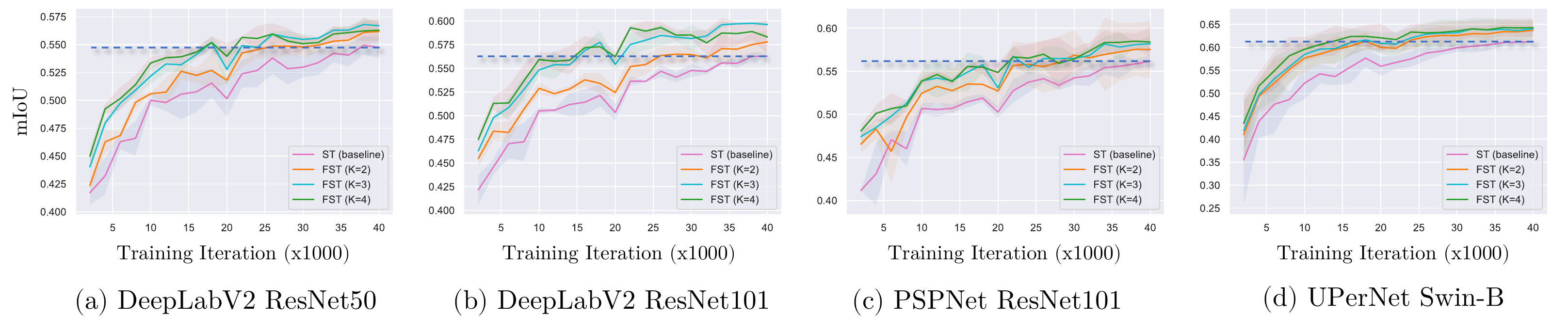}
    \caption{\textbf{Analyses.} Performance curve on validation set during training.}
    \label{fig:sup_miou}
    \vspace{10pt}
    \includegraphics[width=.94\textwidth]{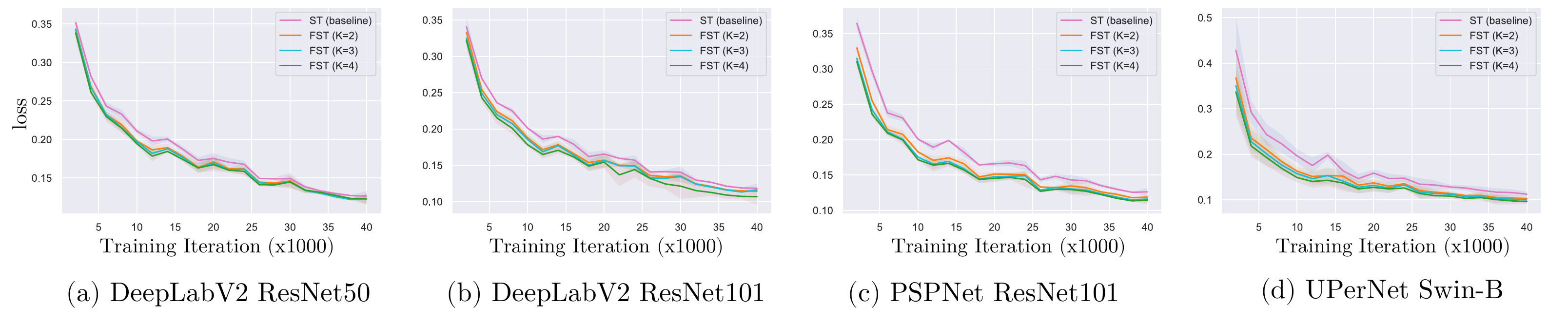}
    \caption{\textbf{Analyses.} Cross-entropy loss on the labeled (training) data during training.}
     \label{fig:sup_loss_labeled}
     \vspace{10pt}
     \includegraphics[width=.94\textwidth]{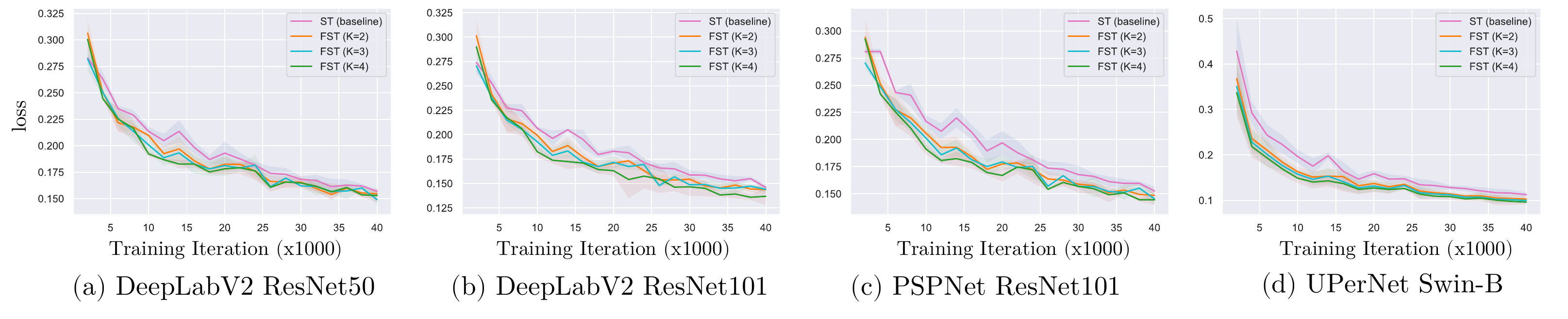}
    \caption{\textbf{Analyses.} Cross-entropy loss on the unlabeled (training) data during training.}
    \label{fig:sup_loss_unlabeled}   
\end{figure}

\begin{figure}[t]
    \centering
    \includegraphics[width=.99\textwidth]{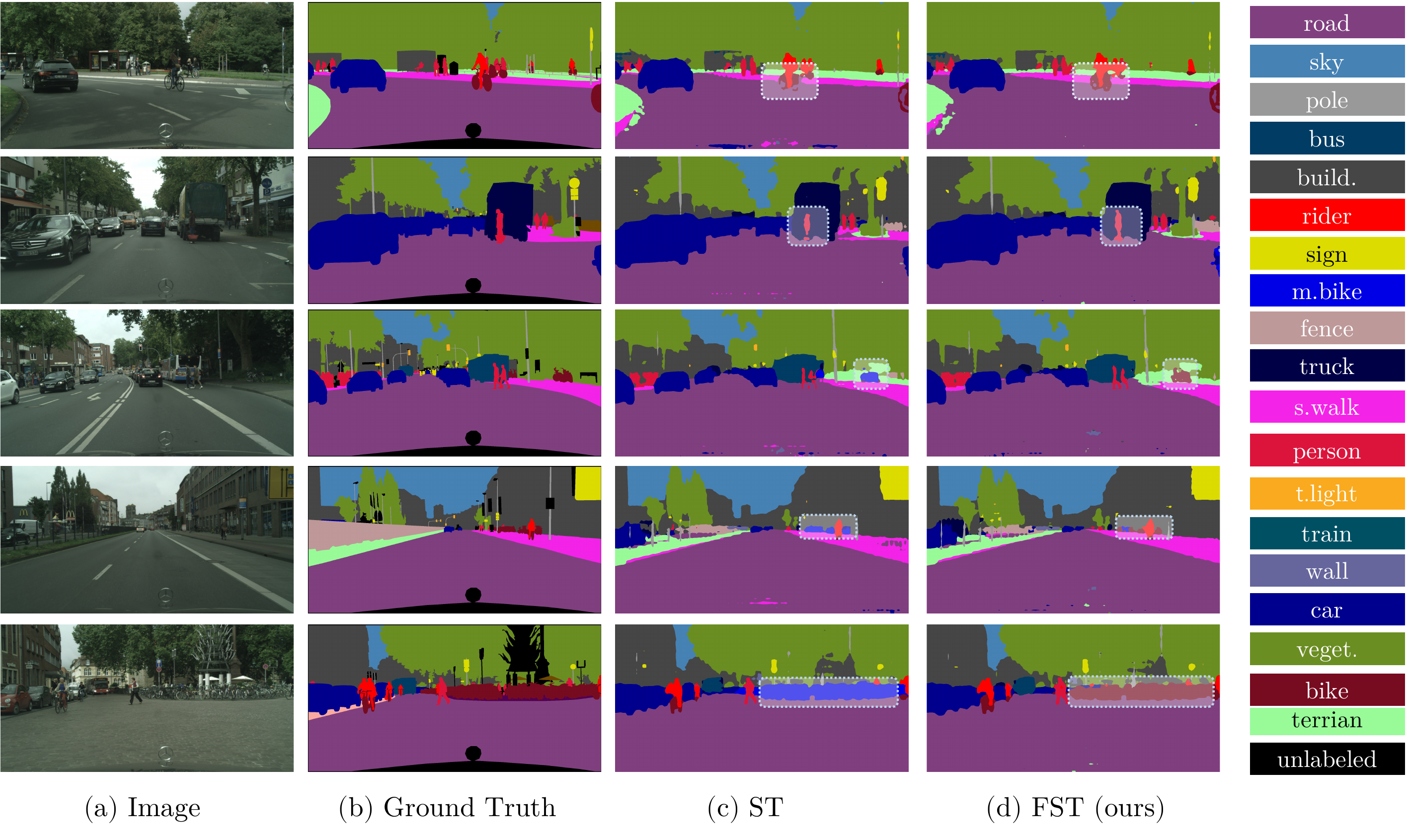}
    \caption{\textbf{Analyses.} 
    More qualitative results on Cityscapes validation set.
    DeepLabV2~\cite{chen2017deeplab} with ResNet-101~\cite{he2016deep} is used.
    }
    \label{fig:sup_more_qualitative_results}
\end{figure}

\cref{fig:sup_miou} presents more performance (mIoU) curves of various network architectures.
We calculate mIoU on the validation set every $2,000$ iterations and plot the mean and standard deviation over $3$ random seeds.
During training, our \method quickly achieves the performance of classical ST, benefiting from the guidance of the estimated future model states.
Moreover, to verify the effect on reducing the confirmation bias, we further observe the training loss on the labeled data (\textit{i.e.}, the training data of source domain), which serves as a complementary to \cref{fig:sup_miou}.
The confirmation bias is considered to mislead the model training.
Here, inspired by~\cite{liu2021cycle}, we empirically observe the bias issue through the model's own training error on the labeled data, since a biased model struggles to fit the labeled samples.
As shown in \cref{fig:sup_loss_labeled}, our \method shows lower cross-entropy loss value of each iteration, especially in the early training stages.
This phenomenon further proves that our \method indeed mitigates the bias problem to some extent.
Note that the presented value in the figure maintains an EMA of the CE loss during training and we plot the mean and standard deviation over $3$ random seeds.
As a comparison, we also plot the training error on the unlabeled data of each iteration, which is shown in \cref{fig:sup_loss_unlabeled}.
Our \method generates higher-quality pseudo-labels on unlabeled samples and achieves lower training error on these samples.
On the one hand, better pseudo-labels make the learning process easier.
On the other hand, due to the mitigation of the confirmation bias, the model reduces the over-fitting to noise pseudo-labels.
Finally, we show more visualization results in \cref{fig:sup_more_qualitative_results} for more qualitative comparisons between ST and our \method.

\section{More discussion} \label{sec-sup:discussion}
\textbf{Broader impact.}
This work mainly focuses on semantic segmentation and its widely adopted momentum teacher-based self-training framework.
However, our approach is a general framework that could be applied to other tasks such as image recognition~\cite{tarvainen2017mean}, object detection~\cite{xu2021end}, few-shot learning~\cite{su2020does} and unsupervised representation learning~\cite{grill2020bootstrap}.
When it comes to other popular online self-training frameworks such as FixMatch~\cite{sohn2020fixmatch}, Noisy student~\cite{xie2020self} and Cycle self-training~\cite{liu2021cycle}, our method is easy to extend by modifying the way of exploiting a model's own future model states.
Besides, our work is compatible with existing appealing technologies such as contrastive learning~\cite{he2020momentum} and active learning~\cite{vezhnevets2012active}. 
We hope our approach can inspire further research about new algorithms, theoretical analyses and applications.

\textbf{Potential negative impact.}
Our work improves the utilization of unlabeled data for semantic segmentation, which could benefit many useful applications such as autonomous driving and remote sensing image analysis.
However, this technology may also be applied to some controversial applications such as surveillance. 
This is a potential risk and a common problem of existing deep learning algorithms and is gaining public attention.
Another possible negative impact is that the learned model could be biased if there was bias in the training data.
Besides, the corresponding carbon emission problem should be considered due to the large-scale data and long-time training of our work.

\section{Pseudo-code} \label{sec-sup:pseudo-code}
To makes our \method easy to understand, we provide pseudo-code in a Pytorch-like style.
To simplify, the improved version of \method (\textit{i.e.}, Eq.~(\textcolor{red}{4})) is implemented in \cref{alg:code}.

\begin{algorithm}[t]
    \caption{Pseudo-code of \method in a PyTorch-like style.}
    \label{alg:code}
    \begin{lstlisting}[language=python]
# g_s, g_t: the student model and the teacher model
# mu, mu': momentum for EMA
# Lambda: dynamic weight to balance the labeled and unlabeled data

g_t.params = g_s.params  # initialize

for (x_l, y_l), x_u in loader:   # load samples
    # momentum update with previous student states
    g_t.params = mu*g_t.params+(1-mu)*g_s.params
    # cache the current student
    g_tmp = g_s.copy()
    # pseudo label prediction: for temp network
    with no_grad():
        y_u = argmax(g_t.forward(x_u))

    # train the temp model
    loss_l = CrossEntropyLoss(g_tmp.forward(x_l), y_l)
    loss_u = CrossEntropyLoss(g_tmp.forward(x_u), y_u)
    loss_virtual = loss_l + Lambda * loss_u   # calculate the loss for temp model
    
    loss_virtual.backward()
    update(g_tmp.params)    # SGD update: temp network
    
    # momentum update with future student states
    g_t.params = mu_prime * g_t.params + (1-mu_prime) * g_tmp.params
    # pseudo label prediction: for student network
    with no_grad():
        y_u = argmax(g_t.forward(x_u))
    
    # train the student
    loss_l = CrossEntropyLoss(g_s.forward(x_l), y_l)
    loss_u = CrossEntropyLoss(g_s.forward(x_u), y_u)
    loss = loss_l + Lambda * loss_u   # calculate loss for student model
    
    loss.backward()
    update(g_s.params)   # SGD update: student network
    
    # delete cache
    del(g_tmp)\end{lstlisting}
\end{algorithm}

\end{document}